\documentclass[aoas]{imsart}

\RequirePackage{amsthm,amsmath,amsfonts,amssymb}
\RequirePackage[authoryear]{natbib}
\RequirePackage[colorlinks,citecolor=blue,urlcolor=blue]{hyperref}
\RequirePackage{graphicx}
\RequirePackage{subcaption}
\RequirePackage{booktabs}
\RequirePackage{multirow}

\startlocaldefs
\theoremstyle{plain}

\newtheorem{remark}{Remark}
\theoremstyle{definition}

\newcommand{\bw}{\boldsymbol{w}}
\newcommand{\bs}{\boldsymbol{s}}
\newcommand{\bx}{\boldsymbol{x}}
\newcommand{\bz}{\boldsymbol{z}}
\newcommand{\bp}{\boldsymbol{p}}

\newcommand{\bI}{\mathbf{I}}
\newcommand{\btheta}{\boldsymbol{\theta}}
\newcommand{\bmu}{\boldsymbol{\mu}}
\newcommand{\bgamma}{\boldsymbol{\gamma}}

\usepackage{xcolor} 
\definecolor{myred}{RGB}{176 48 96}

\endlocaldefs

\begin{document}

\begin{frontmatter}
\title{Syntax-Guided Diffusion Language Models with User-Integrated Personalization}
\runtitle{Syntax-Guided Diffusion Language Models}

\begin{aug}
\author[A]{\fnms{Ruqian}~\snm{Zhang} \ead[label=e1]{rqzhang20@fudan.edu.cn}},
\author[B]{\fnms{Yijiao}~\snm{Zhang}\ead[label=e2]{yijiao.zhang@pennmedicine.upenn.edu}},
\author[A]{\fnms{Juan}~\snm{Shen}\ead[label=e3]{shenjuan@fudan.edu.cn}},
\author[A]{\fnms{Zhongyi}~\snm{Zhu}\ead[label=e4]{zhuzy@fudan.edu.cn}}
\and
\author[C]{\fnms{Annie}~\snm{Qu}\ead[label=e5]{aqu2@ucsb.edu}}
\address[A]{Department of Statistics and Data Science, 
       Fudan University\printead[presep={ ,\ }]{e1,e3,e4}}
\address[B]{Department of Biostatistics, Epidemiology and Informatics, 
       University of Pennsylvania\printead[presep={ ,\ }]{e2}}
\address[C]{Department of Statistics and Applied Probability,
    University of California, Santa Barbara\printead[presep={,\ }]{e5}}
\end{aug}

\begin{abstract}
Large language models have achieved remarkable progress in generating human-like text, yet their outputs often lack structural diversity, limiting personalized expression.
Recent advances in diffusion models offer new opportunities for text generation, surpassing the limitations of autoregressive paradigms.
Building on the idea of hierarchical modeling, we develop a syntax-guided diffusion language model that incorporates latent syntactic variables to enhance text quality, diversity, and personalized conditioning.
The proposed framework decomposes the text distribution into a syntactic prior and a conditional text distribution, enabling interpretable control over both syntactic and lexical components.
We design a cascaded diffusion process for syntax-to-text generation and extend it to a generalized noncascaded formulation with a unified attention mechanism.
To achieve fine-grained personalization, we introduce shared-latent representations that integrate information across users to capture style-specific lexical and syntactic patterns, supporting zero-shot inference.
Extensive experiments demonstrate that the proposed approach consistently improves multiple empirical measures of fluency, diversity, and stylistic fidelity. Further qualitative analyses highlight its interpretability and flexibility in learning personalized patterns.
\end{abstract}

\begin{keyword}
\kwd{Diffusion model}
\kwd{hierarchical generation}
\kwd{natural language processing}
\kwd{representation learning}
\kwd{structural information}
\end{keyword}

\end{frontmatter}

\section{Introduction}
\label{sec:intro}

\begin{figure}[ht]
    \centering
    \includegraphics[width=0.95\linewidth]{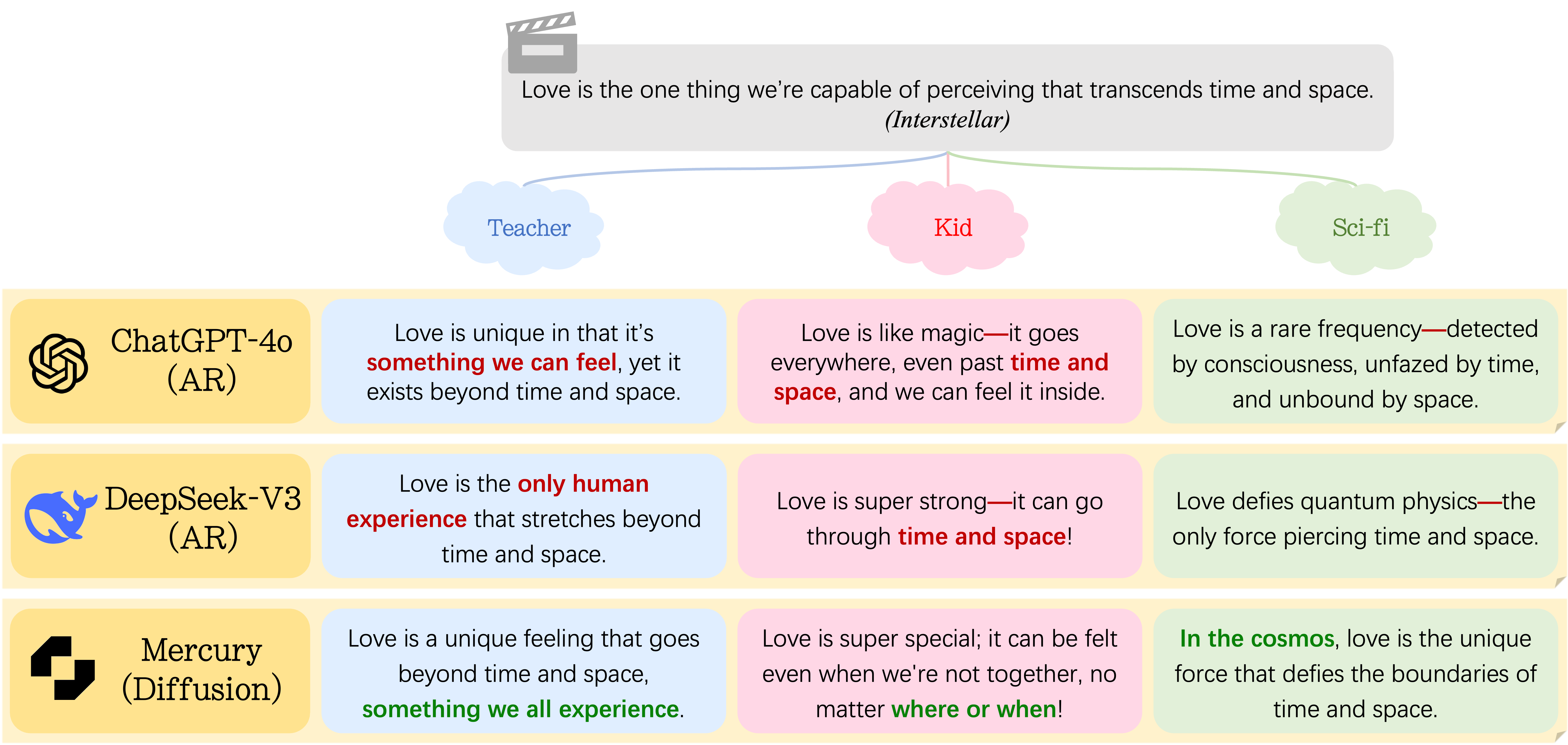}
    \caption{Example of rewriting a movie quote into different styles using autoregressive or diffusion models. Text in red shows limited modifications in words from the AR models, while text in green highlights structural variations introduced by the diffusion approach.}
    \label{fig:prompt-example}
\end{figure}

In recent years, large language models (LLMs) have revolutionized natural language processing (NLP), demonstrating impressive performance in generating human-like text \citep{openai2024gpt4ocard}.
However, their widespread use has also raised growing concerns about linguistic homogenization \citep{wan2023kelly, sourati2025diversity}.
Generated sentences often follow generic templates, leading to overused lexical patterns and repetitive structural compositions, such as the notorious ``em dash (--) conspiracy'' \citep{Mummery368}.
The lack of diversity tends to diminish stylistic richness, posing a challenge for applications demanding personalized expression.
For example, in rewriting a movie quote into three distinct styles using ChatGPT-4o \citep{openai2024gpt4ocard} and DeepSeek-V3 \citep{deepseekai2025deepseekv3}, Figure~\ref{fig:prompt-example} shows that the outputs largely preserve the original structure in the input, with stylistic variations confined mainly to a small set of style-bearing words.

The limited structural diversity of existing LLMs largely arises from their autoregressive (AR) nature, which generates text through a next-token prediction paradigm \citep{Holtzman2020curious}. This left-to-right construction favors high-probability lexical choices as continuations, hindering the ability to refine sentence structure at a global level once tokens are committed.
As a powerful alternative, diffusion-based language models \citep{Li2022LM, nie2025large} are capable of formulating generation as an iterative denoising process \citep{Song2019score, Ho2020DDPM}, and perform parallel sampling across all positions to enable global reorganization of sentence composition.
As illustrated in Figure~\ref{fig:prompt-example}, the diffusion-based Mercury \citep{labs2025mercury} produces greater observed structural and stylistic variation beyond word substitutions.

\begin{figure}[ht]
    \centering
    \begin{subfigure}[b]{0.47\linewidth}
        \centering
        \includegraphics[width=0.9\linewidth]{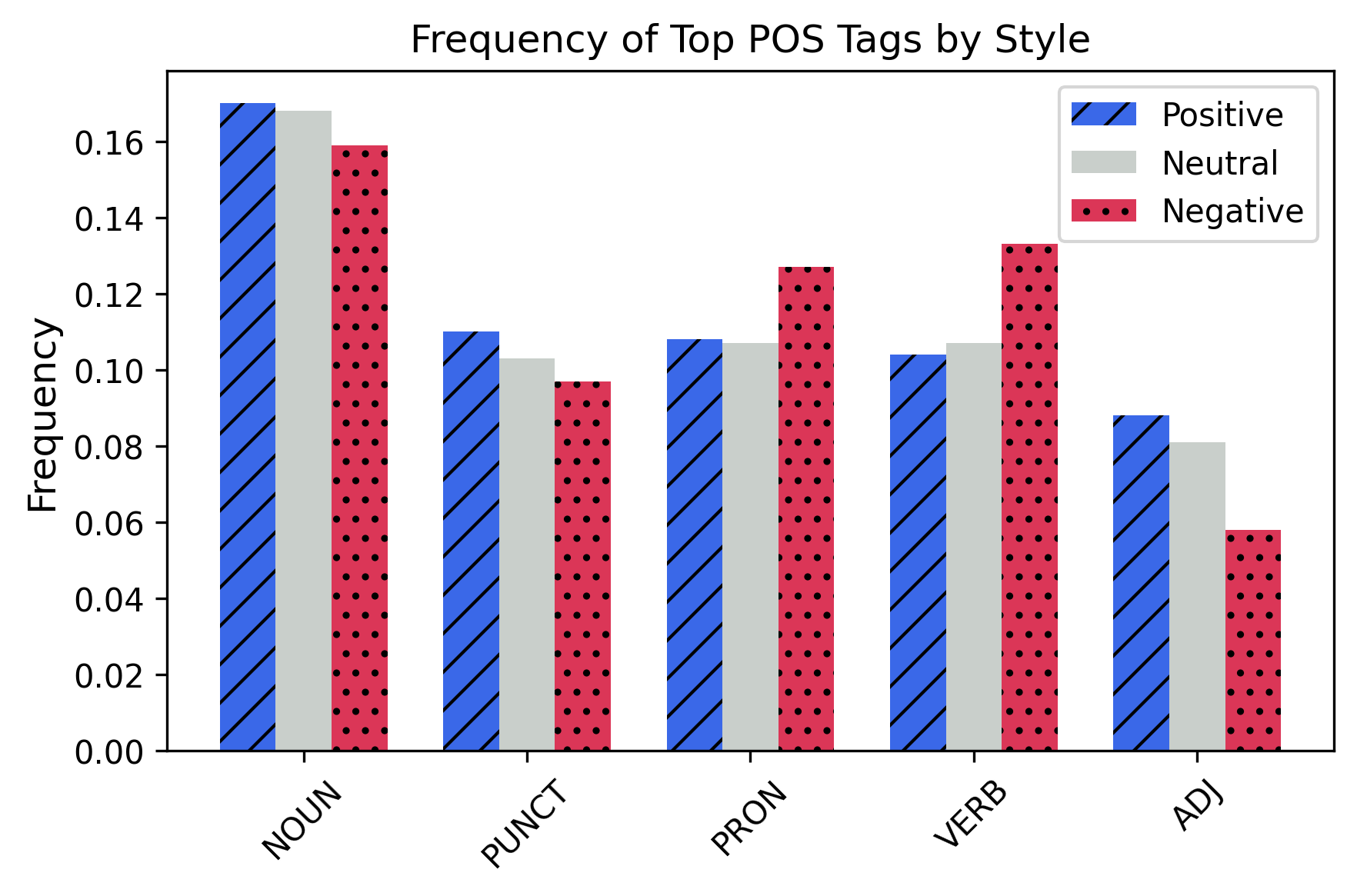}
    \end{subfigure}
    \hfill
    \begin{subfigure}[b]{0.52\linewidth}
        \centering
        \includegraphics[width=\linewidth]{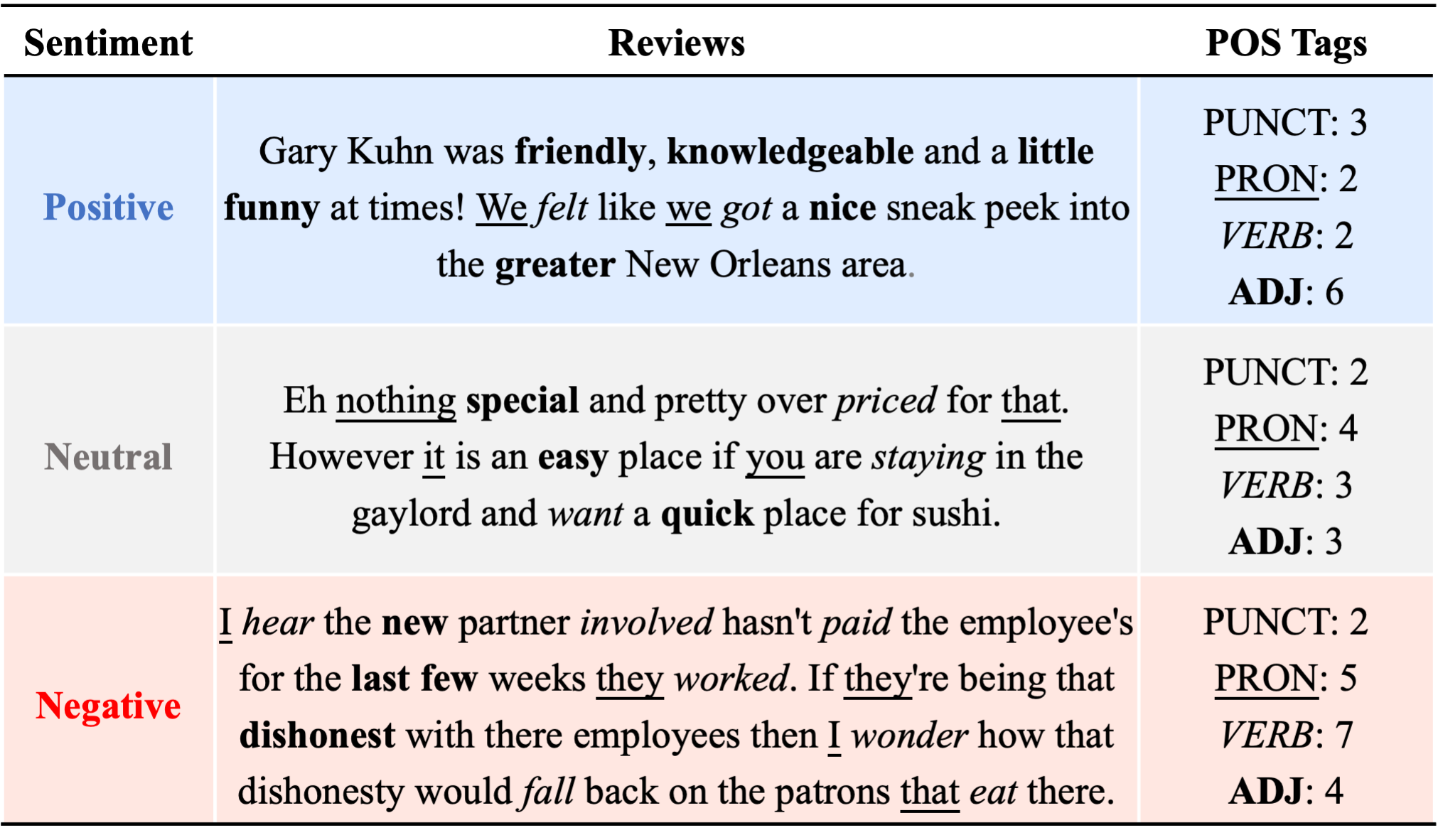}
    \end{subfigure}
    \caption{Left: Frequencies of the most common part-of-speech (POS) tags across sentiment styles in the Yelp Review dataset. Right: Examples of reviews with their POS tags.}
    \label{fig:yelp-pos-freq}
\end{figure}

Motivated by the importance of structural information, we propose a syntax-guided diffusion framework to enhance text diversity and personalized control.
Syntactic patterns offer valuable signals for capturing personalized traits \citep{alhafni2024personalized}.
For example, in the Yelp Review dataset, which comprises user reviews with sentiment labels, part-of-speech (POS) distributions vary across sentiment classes. As shown in Figure~\ref{fig:yelp-pos-freq}, positive reviews favor more frequent use of adjectives (e.g., friendly), while negative reviews use more verbs and pronouns that reflect personal experiences.
Syntax can also contribute to improved text fluency, as it resembles the compositional nature of human language \citep{Chomsky1957}.

Despite the rich information carried by syntax, its use in text generation remains underexplored due to the absence of predefined syntactic inputs in real-world scenarios.
To the best of our knowledge, the application of syntax is limited to sequence-to-sequence tasks with explicit syntactic conditions \citep{li2023Syntactic}.
To fill this gap, motivated by hierarchical generative modeling \citep{blei2003latent,Bachman2016}, we model the text distribution structurally through a marginal syntax distribution and a conditional text distribution given the syntax. 
{Such a decomposition replaces direct learning of the complex marginal text distribution with two more tractable components.
}
Under this formulation, syntactic structures are first drawn and then used as conditions to guide text synthesis in a cascaded pipeline \citep{Ho2022cascade}.
While such cascaded designs have been explored in image generation \citep{ramesh2022unclip, Saharia2022imagen}, their potential for language modeling remains underdeveloped.
Compared with generating text directly from noise, the intermediate syntactic generator serves as a structural prior that may help filter out implausible syntactic patterns and reduce the complexity of lexical selection by narrowing the set of plausible tokens under the generated syntax. 
It also encourages sentence structures aligned with target styles.

\begin{figure}[ht]
    \centering
    \begin{subfigure}[b]{0.50\linewidth}
        \centering
        \includegraphics[width=0.96\linewidth]{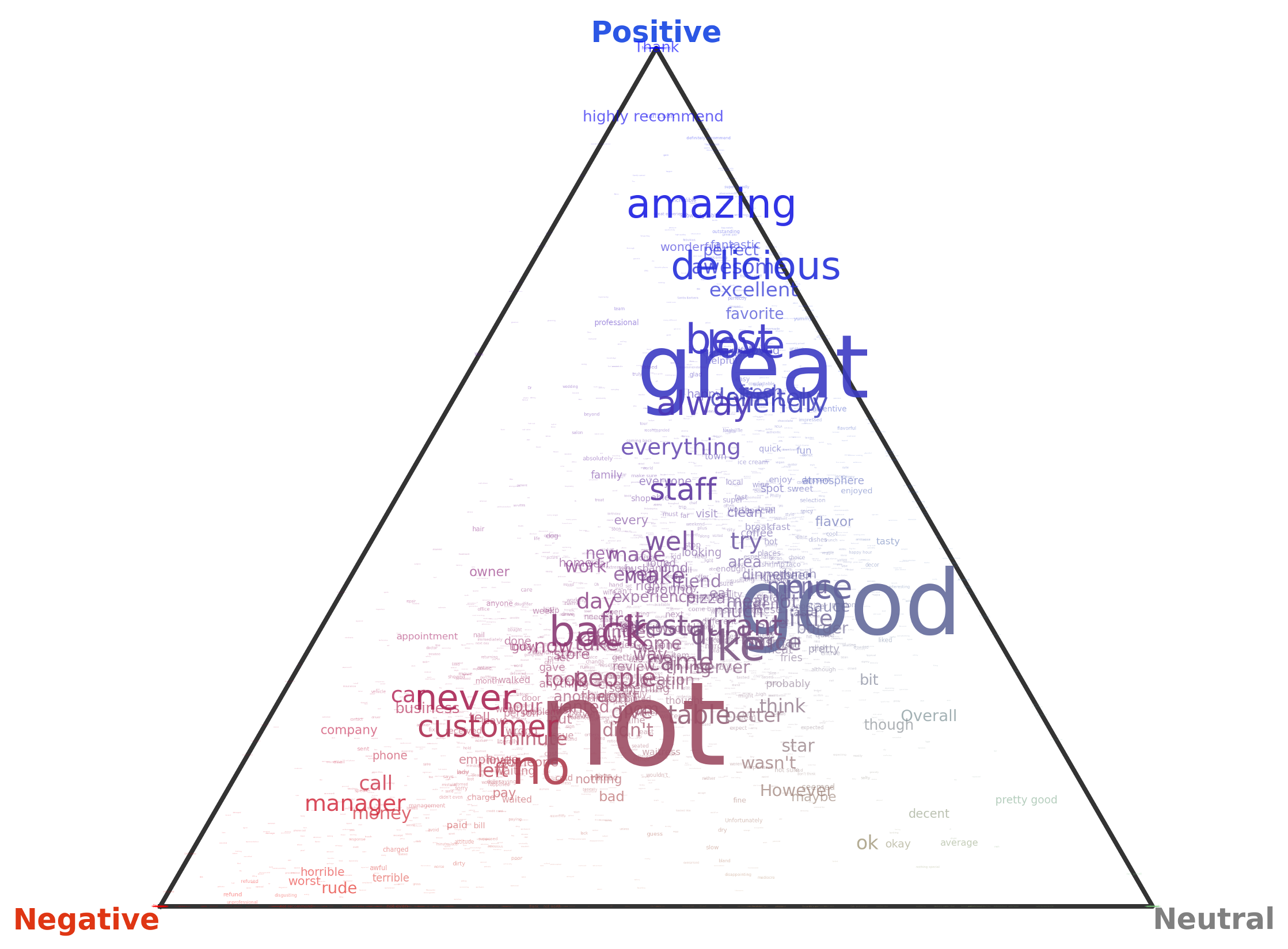}
    \end{subfigure}
    \hfill
    \begin{subfigure}[b]{0.49\linewidth}
        \centering
        \includegraphics[width=0.95\linewidth]{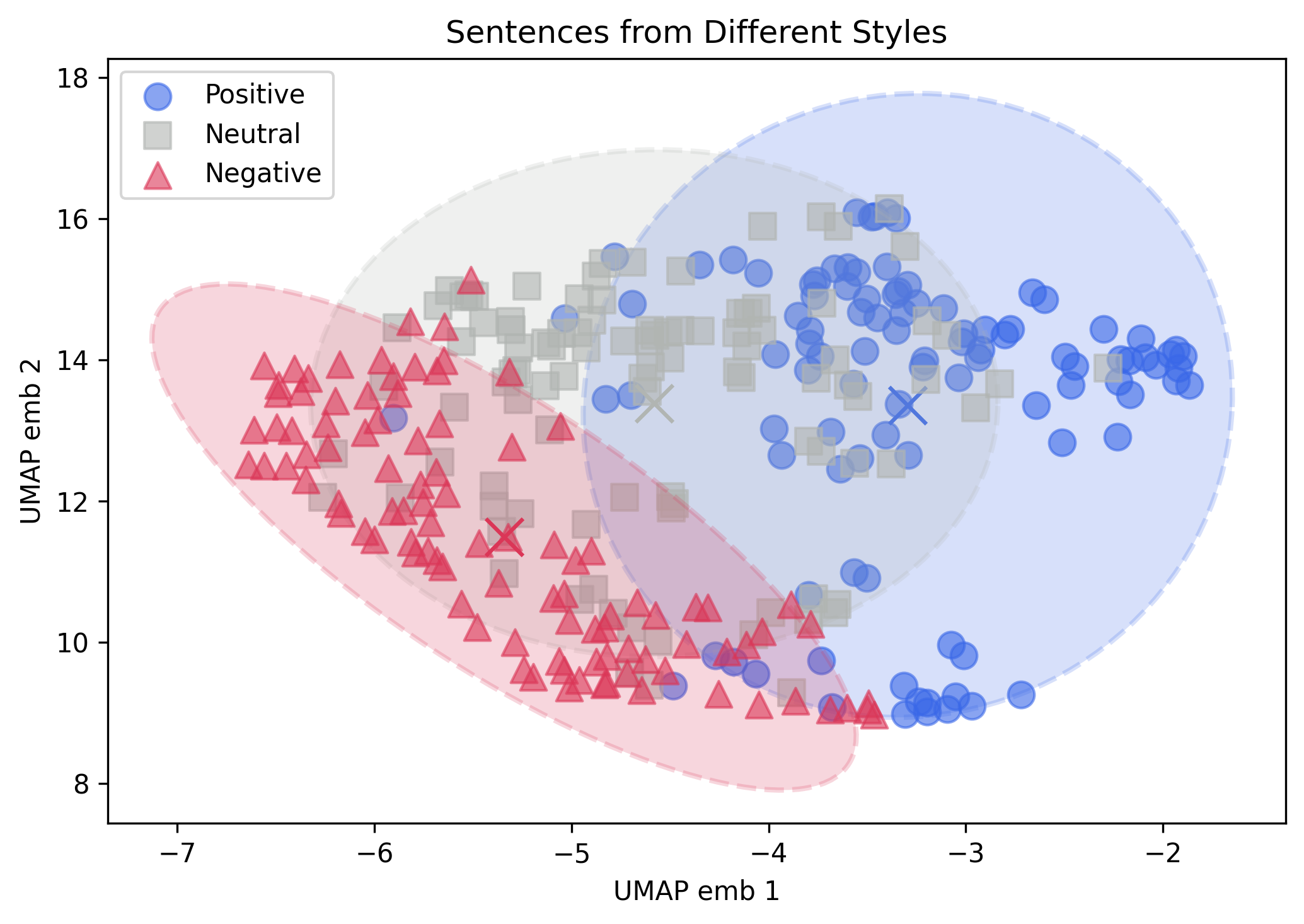}
    \end{subfigure}
    \caption{Left: Word-level distribution in the Yelp Review dataset in a sentiment triangle. Right: Sentence-level UMAP projection of SimCSE embeddings.}
    \label{fig:yelp-text}
\end{figure}

The use of syntax offers a promising direction for personalization.
As shown in Figure~\ref{fig:yelp-pos-freq}, sentiment styles exhibit both distinctive syntactic traits and overlapping patterns. For example, similar frequencies of nouns or punctuations are observed.
However, most existing approaches, such as user-specific identifiers \citep{zhong2021useradapter} or low-rank adaptation \citep{hu2022lora}, typically treat classes in isolation, limiting the potential to share information across styles.
This motivates us to develop a new data integration tool, which is further supported by semantic-level evidence in Figure~\ref{fig:yelp-text}.
Specifically, at the word level, positive reviews tend to use praising words (e.g., great, amazing), while negative ones contain more critical words (e.g., rude, worst).
Neutral reviews share words with both positive and negative styles, while additionally favoring milder vocabulary (e.g., overall, okay).
This interplay extends to the sentence-level representation, illustrated by clustering the UMAP projection \citep{mcinnes2020umap} of SimCSE sentence embeddings \citep{gao2021simcse}, where stylistic boundaries form both distinct and overlapping regions in the latent space.

To better exploit shared and distinctive stylistic signals, we propose a personalization framework based on latent representation learning \citep{xie2020represent,xu2025representation}.
Instead of modeling each style separately, we assume that their personalized embeddings are constructed from a set of building-block personality representations in a shared-latent space, weighted uniquely for each style.
Through integrated modeling, our method leverages all available data to characterize each style, enhancing both generalizability and data efficiency.
To achieve fine-grained control, we move beyond the conventional use of a single embedding vector and introduce an embedding matrix for each style, which interacts with the denoising process through an attention mechanism \citep{Vaswani2017}.
The attention layer is incorporated in both the syntax and text stages of our cascaded diffusion framework.

We further investigate the behavior of multi-stage generation in the cascaded frameworks.
While cascaded diffusion models decompose generation into sequential stages, they often ignore the potential feedback across stages.
To better leverage the parallel nature of diffusion, we propose a generalized noncascaded framework with overlaps between stages, where later predictions can dynamically refine earlier generated conditions through bidirectional interactions.
In addition, we introduce a unified attention mechanism that merges multiple parallel diffusion models into a single, jointly conditional generation process.
By enabling continuous interaction between syntactic and textual components, this unified design improves alignment across stages while reducing overall model complexity.

Our proposed syntax-guided diffusion language model offers a statistical modeling perspective on enhancing the interpretability and stability of generative AI.
Extensive experiments on multiple datasets demonstrate that our method consistently outperforms AR and diffusion baselines of comparable model sizes in both text generation quality and personalized control, validating the effectiveness of syntactic guidance and cross-class information sharing.
We showcase the generalizability of the shared personality representations by extrapolating and interpolating learned class weights to synthesize unseen styles in a zero-shot manner on the Yelp Review dataset.

The remainder of the paper is organized as follows.
{In Section~\ref{sec:cascade}, we present the statistical motivation for syntax-guided distribution decomposition and introduce the cascaded syntax-to-text diffusion framework.}
Section~\ref{sec:person} introduces the integrated personalization method based on shared latent representations.
In Section~\ref{sec:noncascde}, we extend the cascaded framework to a noncascaded formulation that enables bidirectional interactions.
Section~\ref{sec:experiments} reports experimental results, including both quantitative evaluations and illustrative analyses.
Section~\ref{sec:discuss} concludes with potential future directions.
{The implementation code is publicly available at \url{https://github.com/RuqianZhang/syntax-diffusion}.}

\section{Syntax-Guided Text Generation}
\label{sec:cascade}

Text generation aims to model the distribution over text sequences.
Unlike AR models, diffusion language models jointly learn the distribution of all tokens in a sequence, offering higher diversity and flexibility in text synthesis \citep{Hoogeboom2021, chen2023analog, MDLM}.
A major line of work operates in continuous embedding spaces \citep{Li2022LM}, which has proven effective in improving efficiency and quality \citep{Lovelace2023latent}.
While diffusion language models have shown strong potential in LLM \citep{Li2023survey}, they remain in an early stage of development and often underperform AR counterparts in terms of text quality \citep{gong2023diffuseq, rout2025anchored}, underscoring the need for improvement.

\subsection{Statistical Motivation of Syntactic Guidance}

Given a text sequence $\bw_x$ of length $L$, we denote its embeddings as 
$\bx_0 = E_x(\bw_x) \in \mathbb{R}^{L \times d}$, where $E_x$ is a pretrained text encoder and $d$ is the embedding dimension.
The marginal distribution of human text $q(\bx_0)$ is highly complicated, since it simultaneously mixes multiple sources of variation, including structural pattern, lexical choice, and semantic content.
Directly learning such a distribution through a single model $p_{\btheta}(\bx_0)$ can therefore be statistically and computationally challenging.

To make the learning problem more manageable, we consider data augmentation by introducing structural variables $\bz$ and decomposing the marginal text distribution as
\begin{equation}
\label{eq:dist_decomp}
	q(\bx_0)=\int q(\bx_0 \mid \bz )q(\bz)d\bz,
\end{equation}
where $q(\bz)$ represents the structural distribution and $q(\bx_0\mid\bz)$ represents the conditional text distribution given the structure.
Building on this formulation, instead of directly modeling the marginal text distribution, we model the generation process through two components,
\begin{equation}
\label{eq:model_dist_decomp}
    p_{\btheta}(\bx_0)
    =
    \int p_{\btheta_x}(\bx_0\mid\bz)p_{\btheta_z}(\bz)d\bz.
\end{equation}
This decomposition separates the modeling problem into two components: the structural model $p_{\btheta_z}(\bz)$ and the conditional text model $p_{\btheta_x}(\bx_0\mid\bz)$. 
The benefit of decomposition \eqref{eq:dist_decomp} depends on two requirements:
\begin{enumerate}
    \item[(a)] The structural distribution $q(\bz)$ should be simpler than the full text distribution $q(\bx_0)$;
    \item[(b)] The structure $\bz$ should explain a meaningful portion of variation in $\bx_0$, so that $q(\bx_0\mid\bz)$ is simpler and more concentrated than the original marginal distribution.
\end{enumerate}
This viewpoint is related in spirit to latent factors or low-rank decompositions, where a complicated distribution is represented through lower-complexity components, thereby reducing the effective dimension of the learning problem \citep{Blei2014review, Noirrit2023latent, Chen01122012, Fan2013lowrank}.

Syntax provides a natural and interpretable choice for structural information in language analysis.
Unlike images, whose structure is often encoded implicitly in a latent space \citep{hong2018inferring}, syntax is explicitly observable in text, allowing for linguistically grounded guidance.
Syntax satisfies both requirements naturally: (a) it is substantially simpler than text with a small vocabulary size; and (b) it induces a partition of the text space into structurally more homogeneous subsets.
Once the syntactic category at a position is specified, the set of plausible lexical choices is sharply reduced, making the conditional distribution $q(\bx_0\mid\bz)$ easier to model.
Moreover, syntax provides an additional source of supervision, which is particularly valuable when human-written text is scarce.
In this work, we adopt part-of-speech (POS) tags as the structural representation, which can be easily extracted using tools such as spaCy \citep{spacy2}.
Details of POS tags are provided in Section C.1 of the Supplementary Material.

\subsection{Cascaded Syntax-to-Text Diffusion}

\begin{figure}[ht]
    \centering
    \includegraphics[width=\textwidth]{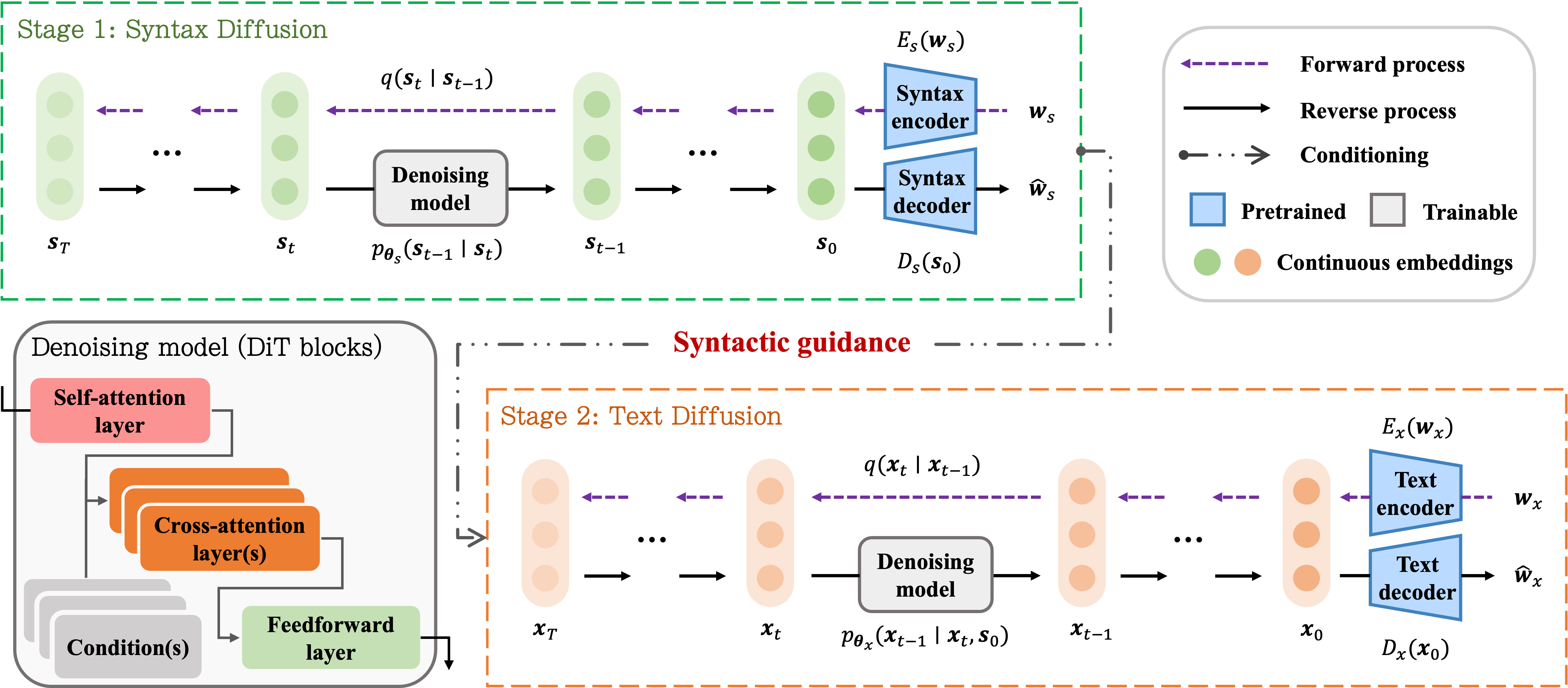}
    \caption{Overview of the proposed cascaded syntax-to-text diffusion. The model first generates syntactic embeddings through a syntax diffusion model and then generates text embeddings through a text diffusion model conditioned on the generated syntax.
    The cross-attention layers in the denoising models incorporate stage-specific and task-specific conditions, including syntactic guidance, personalization information, and contextual prompts. The integration of personalization into the syntax and text diffusion stages is detailed in Section~\ref{sec:player}.}
    \label{fig:overview}
\end{figure}

Building on the formulation \eqref{eq:model_dist_decomp}, we propose a hierarchical diffusion framework that decomposes text generation into two cascaded stages: first generating syntax as a structural prior, and then generating text conditioned on the generated syntax.
We refer to this cascaded syntax-to-text diffusion as \textbf{SynText}. An overview is shown in Figure~\ref{fig:overview}.

\subsubsection{Learning to Generate Syntax as Structural Guidance}
\label{sec:syn_diff}

Given a text sequence $\bw_x$, we extract its POS sequence $\bw_s$.
To model syntactic structure within a diffusion framework, we adapt the approach in \citet{Li2022LM}, using a pretrained syntactic encoder $E_s$ to project $\bw_s$ into a continuous latent space and yield syntactic embeddings $\bs_0 = E_s(\bw_s) \in \mathbb{R}^{L \times d}$.

We denote the true data-generating distribution of the syntactic embeddings as $\bs_0\sim q(\bs_0)$.
The forward diffusion process progressively perturbs $\bs_0$ over $T$ timesteps by adding Gaussian noises, governed by a predefined variance schedule $\{\alpha^{s}_t\}_{t=1}^T$ with $\alpha^{s}_t\in(0,1)$:
\begin{equation*} 
	q(\bs_t \mid \bs_{t-1}) = \mathcal{N}(\sqrt{{\alpha}^{s}_t} \bs_{t-1}, (1 - \alpha^{s}_t) \bI),
\end{equation*}
where $\bs_{t}$ denotes the noisy syntactic embeddings at time $t$ and $\bI$ is the identity matrix.
This Markovian corruption process produces a sequence of latent variables $\{\bs_t\}_{t=1}^T$ with decreasing signal-to-noise ratio. As $T$ becomes sufficiently large, the marginal distribution of $\bs_T$ converges to a standard multivariate Gaussian distribution $\mathcal{N}(\boldsymbol{0}, \bI)$.
If the reverse distribution $q(\bs_{t-1}\mid \bs_t)$ were available, we could generate new samples from $q(\bs_0)$ by first sampling $\bs_T\sim \mathcal{N}(0, \bI)$ and then iteratively removing noises.

Since $q(\bs_{t-1}\mid \bs_t)$ is computationally intractable, we adopt variational approximation with a reverse model $p_{\btheta_s}(\bs_{t-1}\mid \bs_{t})$, where $\btheta_s$ denotes the trainable parameters.
This reverse process learns to denoise the latent sequence to reconstruct samples from noise.
To train the model, we minimize the KL divergence between the true joint distribution $q(\bs_{0:T})$ and the parametrized joint distribution $p_{\btheta_s}(\bs_{0:T})$, which is equivalent to maximizing a variational lower bound on the log-likelihood $\log p_{\btheta_s}(\bs_0)$. The objective function is given by
\begin{equation}
\label{eq:syn_loss}
\begin{aligned}
    {\mathcal L}_{s}(\btheta_s)&\equiv D_{\operatorname{KL}}(q(\bs_{0:T})\| p_{\btheta_s}(\bs_{0:T}))
    =\mathbb{E}_{q(\bs_{0:T})}\left[\log \frac{q(\bs_{0:T})}{p_{\btheta_s}(\bs_{0:T})} \right]\\
    &=\mathbb{E}_{q(\bs_{0:T})} \left[ \log \frac{q(\bs_T \mid \bs_0)}{p_{\btheta_s}(\bs_T)}
    + \sum_{t=2}^{T} \log \frac{q(\bs_{t-1} \mid \bs_t, \bs_0)}{p_{\btheta_s}(\bs_{t-1} \mid \bs_t)}
    + \log \frac{q(\bs_0)}{p_{\btheta_s}(\bs_0 \mid \bs_1)} \right],
\end{aligned}
\end{equation}
where $p_{\btheta_s}(\bs_T)={\mathcal N}(0,\bI)$ serves as the prior. Following \citet{Ho2020DDPM}, we model the reverse transition $p_{\btheta_s}(\bs_{t-1}\mid \bs_{t})$ as a Gaussian distribution with fixed variance:
\begin{equation*}
    p_{\btheta_s}(\bs_{t-1}\mid \bs_{t})={\mathcal N}(\bmu_{\btheta_s}(\bs_t,t), (1-\alpha^{s}_t)\bI),
\end{equation*}
where $\bmu_{\btheta_s}(\bs_t,t)$ is predicted by a neural network, implemented using a Transformer-based architecture \citep{Vaswani2017, Peebles2023DiT}. This parameterization mirrors the forward process and allows efficient optimization of the objective.

Leveraging the syntactic diffusion model, we can generate syntactic conditions $\hat{\bw}_s$ from scratch by first sampling random noises $\hat{\bs}_T\sim{\mathcal N}(\boldsymbol{0},\bI)$ and iteratively denoising through the learned reverse model $p_{\btheta_s}(\hat{\bs}_{t-1}\mid \hat{\bs}_{t})$ for $t=T, T-1,\ldots,1$.
The final denoised embedding $\hat{\bs}_0$ is reconstructed into a POS sequence using a pretrained syntactic decoder $D_s$.
The syntactic diffusion model forms the first stage of our hierarchical generation pipeline, enabling the synthesis of structural conditions without predefined syntactic inputs, thereby allowing more flexible downstream text generation.

\subsubsection{Cascaded Text Generation with Syntactic Conditioning}
\label{sec:text_diff}

The cascaded text generation leverages syntactic structure through a two-stage process as $p_{\btheta_x}(\bx_0 \mid \bs_0)p_{\btheta_s}(\bs_0)$ with $\btheta_x$ denoting the trainable parameters of the text diffusion model.
In the first stage, $p_{\btheta_s}(\bs_0)$ serves as a prior over syntactic structure, as described in Section~\ref{sec:syn_diff}.
In the second stage, $p_{\btheta_x}(\bx_0 \mid \bs_0)$ models the conditional distribution over text given the prior, formulating the generation task as a sequence-to-sequence problem.

To model the second stage, we introduce a conditional text diffusion model that incorporates syntactic information via an attention mechanism.
The conditional denoising model is defined as $p_{\btheta_x}(\bx_{t-1}\mid \bx_{t}, \bs_0)={\mathcal N}(\bmu_{\btheta_x}(\bx_t, \bs_0,t), (1-\alpha^{x}_t)\bI)$ with a schedule $\{\alpha^{x}_t\}^T_{t=1}$, which computes cross-attention between the noisy text embeddings $\bx_t$ and $\bs_0$ at $t$. Specifically, we define learnable projection matrices for the query, key, and value as $W_Q,W_K,W_V\in\mathbb{R}^{d\times d_M}$, where $d_M$ is the dimension of the projected space.
The query $Q(\bx_t)$ is computed as $Q(\bx_t)=\bx_t W_Q$, while the key and value are $K(\bs_0)=\bs_0 W_K$ and $V(\bs_0)=\bs_0 W_V$, respectively.
The attention map is then given by $A=\operatorname{Softmax}(Q(\bx_t)K(\bs_0)^T/{\sqrt{d_M}})$,
which captures the similarity between each query and key token. The output of the attention layer is a weighted combination of the value vectors as $AV(\bs_0)$.

For the training of the text reverse model, we define a forward diffusion process analogous to that in Section~\ref{sec:syn_diff}, where the text embeddings $\bx_0$ are progressively corrupted over $T$ steps according to $q(\bx_t \mid \bx_{t-1}) = \mathcal{N}(\sqrt{{\alpha}^{x}_t} \bx_{t-1}, (1-\alpha^{x}_t) \bI)$.
The model parameters $\btheta_x$ are learned by minimizing the objective function similar to (\ref{eq:syn_loss}) with $p_{\btheta_x}(\bx_T)={\mathcal N}(0,\bI)$:
\begin{equation*}
\label{eq:text_loss}
\begin{aligned}
    {\mathcal L}_{x}(\btheta_x)
    =\mathbb{E}_{q(\bx_{0:T})} \left[ \log \frac{q(\bx_T \mid\bx_0)}{p_{\btheta_x}(\bx_T)}
    + \sum_{t=2}^{T} \log \frac{q(\bx_{t-1} \mid \bx_t, \bx_0)}{p_{\btheta_x}(\bx_{t-1} \mid \bx_t, \bs_0)}
    + \log \frac{q(\bx_0)}{p_{\btheta_x}(\bx_0 \mid \bx_1, \bs_0)} \right].
\end{aligned}
\end{equation*}

The syntactic and text diffusion models jointly form the cascaded generation pipeline.
To generate a new sentence, the process begins with the syntactic diffusion model, which transforms a noise sample $\hat{\bs}_T\sim{\mathcal N}(\boldsymbol{0}, \bI)$ into syntactic embeddings $\hat{\bs}_0$ via iteratively denoising.
These embeddings then guide the text diffusion model, which denoises a second random noise sample $\hat{\bx}_T\sim{\mathcal N}(\boldsymbol{0}, \bI)$ to produce the text embeddings $\hat{\bx}_0$.
The output sentence $\hat{\bw}_x$ is reconstructed from $\hat{\bx}_0$ using the text decoder $D_x$.
This two-stage framework enriches learnable information and enhances generation quality by explicitly modeling syntactic structure before text realization.
The detailed training and generation procedures are summarized in Algorithms 1 and 2 in Section B.1 of the Supplementary Material.

Our proposed SynText exploits the potential of syntax as structural guidance in text generation.
Beyond improving quality and diversity, syntax lays a new foundation for fine-grained personalization through stylistic structural patterns.
Syntax also offers greater interpretability from a linguistic perspective, providing reasoning into sentence structure beyond the probabilistic predictions of latent diffusion models.
Nevertheless, the cascaded framework comes with limitations. It requires an additional syntactic diffusion model, and the two-stage design imposes a one-way direction of dependency between syntax and text, which may induce error propagation if syntactic predictions are imperfect.
These challenges motivate us to develop more flexible and efficient architectures, which we address in Section~\ref{sec:noncascde}.

\section{Personalization with Shared Personalities}
\label{sec:person}

\subsection{Shared Latent Personality Representations}

Personalized text generation seeks to align outputs with specific preferences by capturing stylistic attributes \citep{salemi2024lamp}.
Existing methods typically treat each style in isolation, learning separate parameters to model stylistic variations.
For example, unique identifier tokens are introduced to encode user information into latent embeddings \citep{zhong2021useradapter, mireshghallah2022useridentifier},
or user-specific adapters, such as low-rank adaptation, are adopted to tune selected layers \citep{hu2022lora, hayou24lora}.
While effective, these class-by-class approaches restrict information integration across styles, leading to inefficient data utilization and limited generalization.
Furthermore, the complexity of model architectures requires careful layer selection in LoRA tuning \citep{frenkel2024implicit}, undermining its interpretability.

\begin{figure}[ht]
    \centering
    \includegraphics[width=0.6\linewidth]{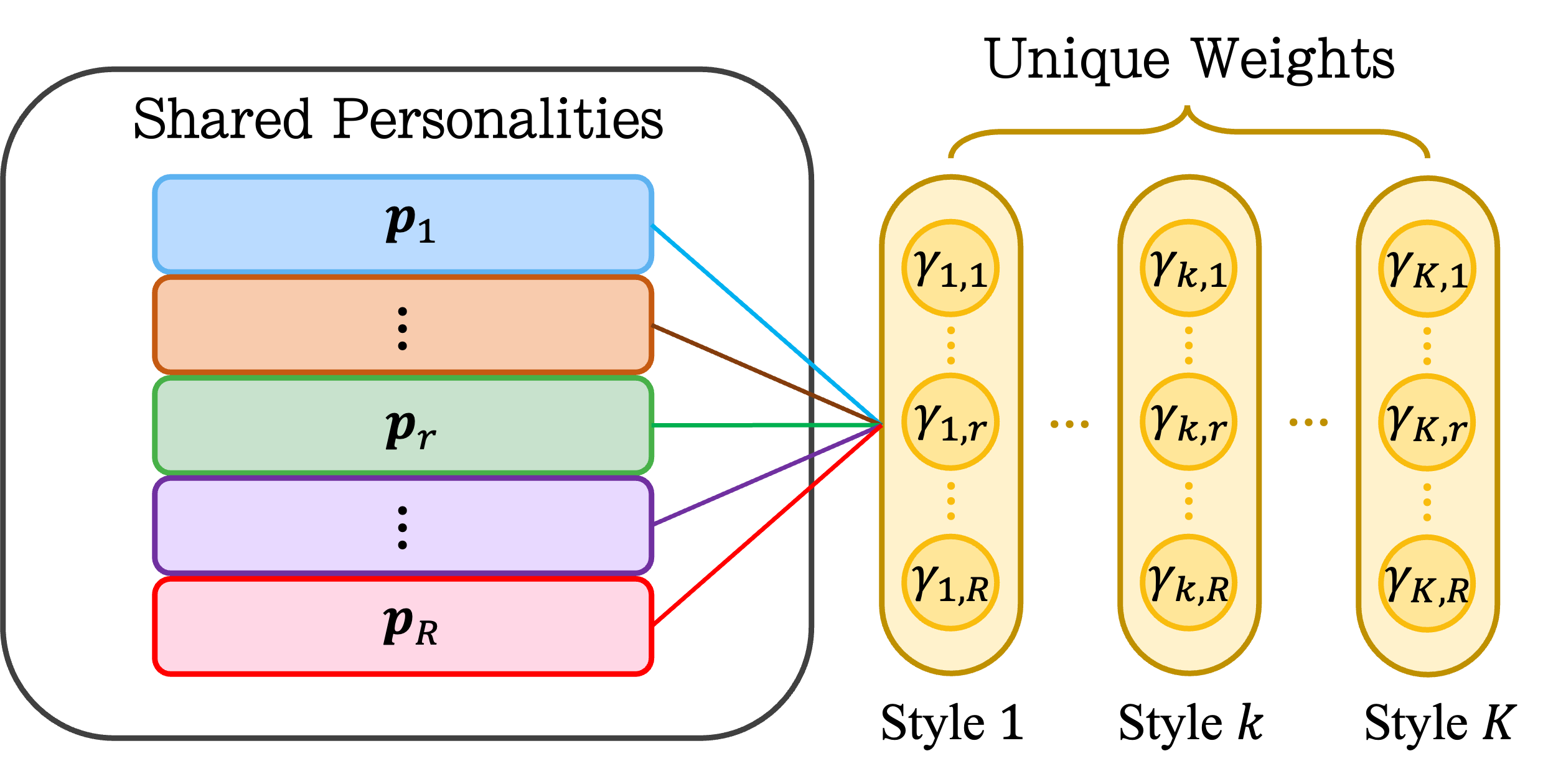}
    \caption{Illustration of shared personality representations and unique personality weights.}
    \label{fig:embedding_matrix}
\end{figure}

To address these challenges, we propose a flexible method based on information sharing and incorporate it into both the syntax and text diffusion stages. 
This design is closely related to statistical representation learning and multi-task learning, where related tasks are jointly learned through shared latent representations while retaining task-specific components \citep{xie2020represent, Collins2024multi, Sui2025multi}.
Motivated by the observations in Figures~\ref{fig:yelp-pos-freq} and \ref{fig:yelp-text}, we assume the existence of a shared representation space across all styles. We construct this space by introducing a set of shared latent representations, termed \textit{personalities}, which serve as building blocks to unify style-specific embeddings.
Specifically, assume there are $R$ common personality features, denoted as $\bp_1,\ldots,\bp_R\in \mathbb{R}^{d_p}$, where $d_p$ is the embedding dimension.
We define the personality matrix as $\boldsymbol{P}=(\bp_1,\ldots,\bp_R)\in\mathbb{R}^{d_p\times R}$, which serves as a codebook storing all personalized factors shared across styles.
Suppose the dataset contains $K$ different styles.
For each style $k\in\{1,\ldots,K\}$, we introduce a style-specific trainable weight vector 
$\bgamma_k=(\gamma_{k,1},\ldots,\gamma_{k,R})^T$.
Each entry $\gamma_{k,r}\in[0,1]$ represents how much the $r\rm{th}$ personality is expressed in style $k$ and satisfies $\sum^R_{r=1}\gamma_{k,r}=1$.
The personalized embeddings are then constructed by extracting relevant personalities in $\boldsymbol{P}$ according to a personalized weight vector $\bgamma_k$, as illustrated in Figure~\ref{fig:embedding_matrix}.

Our framework decomposes personalization into two components: a shared set of personality representations and distinct personalized weights, enabling better integration of cross-style information.
By avoiding the need to learn separate embeddings for each style, the model significantly reduces the number of style-specific parameters from $d_p$ to $R$, alleviating the reliance on per-style data and improving sample efficiency.
The introduction of personality weights also enhances interpretability, which characterizes the composition of each style and associations across styles.

The unified personality space facilitates generalization to unseen styles. 
Once the shared personality representations are learned, they span a latent stylistic space in which new styles can be synthesized by adjusting the weights over the observed styles. 
Specifically, for an unseen style, we can construct a new weight vector by interpolating or extrapolating the learned style-specific weights $\{\bgamma_k\}_{k=1}^K$, enabling zero-shot generation as illustrated in Section~\ref{sec:qual_illu}.

\subsection{Personality Layer with Attention Mechanism}
\label{sec:player}
While personalized embeddings are typically defined as vectors from convex combinations of personality representations, $\boldsymbol{e}_k=\boldsymbol{P}\bgamma_k=\sum^{R}_{r=1}\gamma_{k,r}\bp_r$, we extend this approach to a matrix form $\boldsymbol{E}_k=[\gamma_{k,1}\bp_1,\ldots,\gamma_{k,R}\bp_R]^T\in\mathbb{R}^{R\times d_p}$ for $k=1,\ldots, K$.
Unlike conventional methods that concatenate a fixed personalized token into the noisy input during the denoising process, our matrix-based design simulates a personalized prompt of length $R$, enabling the model to attend over multiple personality components through a cross-attention mechanism.
This formulation facilitates dynamic and context-aware personalized conditioning during generation, offering greater flexibility.

\begin{figure}[ht]
    \centering
    \includegraphics[width=\linewidth]{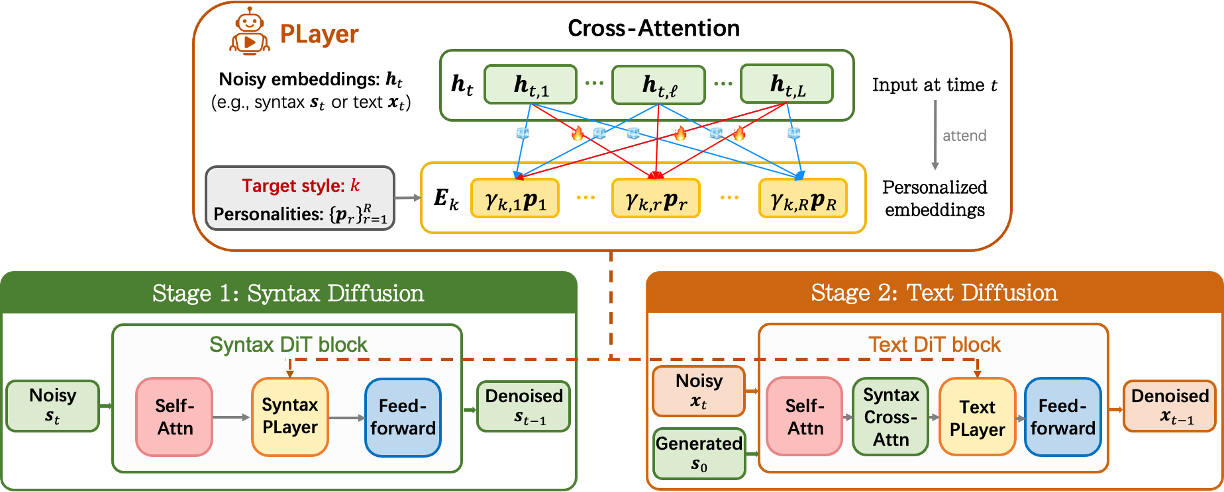}
    \caption{
    Illustration of the generic cross-attention mechanism used by PLayer for personalization and its stage-specific integration into the cascaded syntax-to-text diffusion framework.
    }
    \label{fig:Player}
\end{figure}

When applied in diffusion models, the personalized attention mechanism introduces a personality layer in the Transformer block, referred to as \textbf{PLayer}. For a given style $k$, we compute the key and value from $\boldsymbol{E}_k$ as $K(\boldsymbol{E}_k) = \boldsymbol{E}_k W_K$ and $V(\boldsymbol{E}_k) = \boldsymbol{E}_k W_V$, where $W_K, W_V\in \mathbb{R}^{d_p\times d_M}$ are projection matrices. 
At each denoising timestep $t$, the query derived from the noisy embeddings $\boldsymbol{h}_t$ is denoted as $Q(\boldsymbol{h}_t) \in \mathbb{R}^{L \times d_M}$, where $\boldsymbol{h}_t=\bs_t$ for syntax diffusion and $\boldsymbol{h}_t=\bx_t$ for text diffusion.
The generation is then guided toward the desired stylistic direction through cross-attention as illustrated in Figure~\ref{fig:Player}.

In our cascaded framework, stage-specific PLayer modules are incorporated into the syntax and text diffusion stages, enabling the model to capture richer stylistic signals than traditional diffusion language models that condition only on the text level.
In the syntax stage, the target style conditions syntax generation through the syntax PLayer.
In the text stage, two separate cross-attention layers provide syntactic guidance and personalized conditioning, respectively.
The two PLayer modules have independent parameters.
During training, the same style label is provided to both the syntax diffusion model and the text diffusion model, and each PLayer module is optimized jointly with its corresponding denoising model.
Implementation details of PLayer are provided in Section B.3 of the Supplementary Material.

\section{Noncascaded Generation with Mutual Guidance}
\label{sec:noncascde}

\subsection{From Cascade to Noncascade}

\begin{figure}[ht]
    \centering
    \begin{subfigure}{0.48\textwidth}
    \centering
        \includegraphics[width=0.8\linewidth]{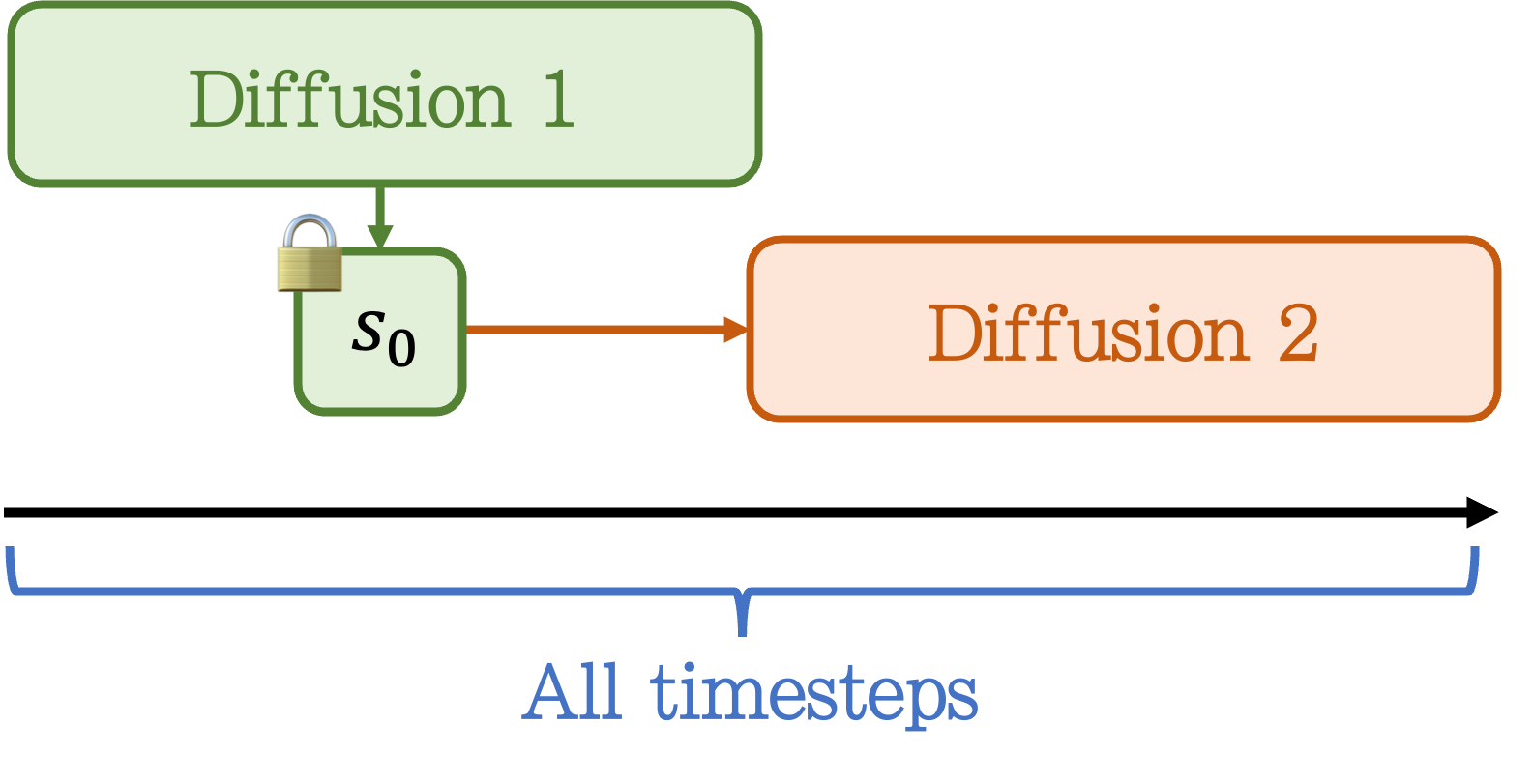}
        \caption{Cascade}
        \label{fig:cascade}
    \end{subfigure}
    \hfill 
    \begin{subfigure}{0.51\textwidth}
    \centering
        \includegraphics[width=0.8\linewidth]{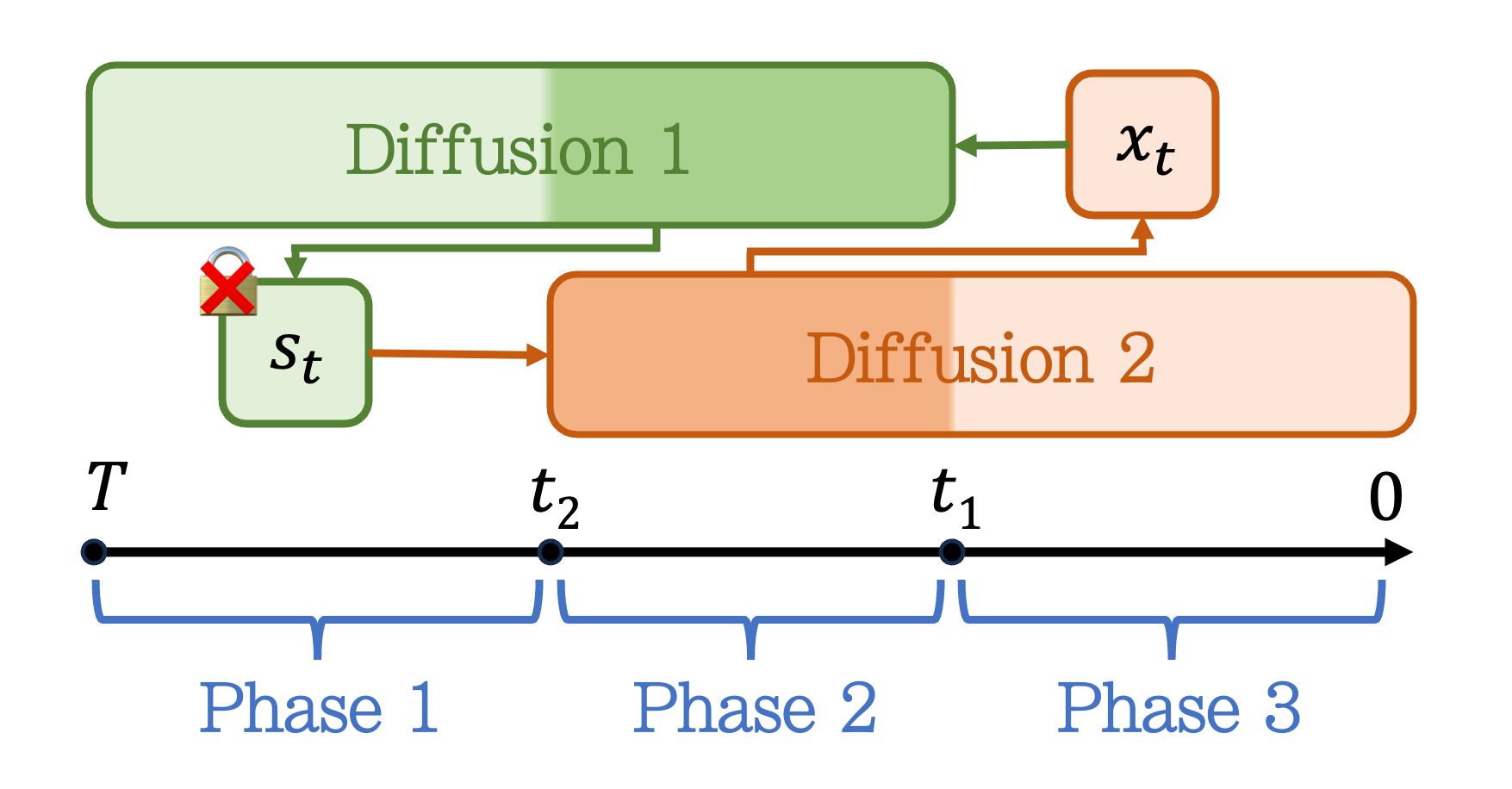}
        \caption{Noncascade}
        \label{fig:overlap}
    \end{subfigure}
     \caption{Generating processes of the cascaded diffusion and the noncascaded diffusion.}
    \label{fig:cas_non}
\end{figure}

The cascaded framework defines a sequential order for generating multiple outputs, where each stage begins after the previous one has fully produced its results, as illustrated in Figure~\ref{fig:cas_non}(\subref{fig:cascade}).
For example, in \textbf{SynText}, syntactic structures are first generated and then used to guide subsequent text realization.
However, the information flow between structure and content can be bidirectional, where semantics may also shape syntactic form. 

To accommodate bidirectional interactions, we propose a noncascaded diffusion framework by introducing temporal overlap between denoising processes, as shown in Figure~\ref{fig:cas_non}(\subref{fig:overlap}).
Instead of obtaining a fixed condition from the first stage, the two diffusion models proceed concurrently during the overlap phase, allowing dynamic information exchange from one to the other.
The intermediate predictions from the second diffusion serve as real-time feedback to the first, promoting mutual adaptation throughout the generation.

Specifically, consider the two-stage case of syntactic and text diffusion models. Introducing temporal overlap divides the generation process into three phases. 
While the forward processes remain identical to those in Section~\ref{sec:cascade}, the reverse processes are redesigned to allow mutual dependency between modalities.
In \textit{Phase 1}, from $T$ to $t_2$, the model focuses on syntax, denoising random noise into intermediate syntactic embeddings by $p_{\btheta_s}(\bs_{t-1}\mid \bs_{t})$.
In \textit{Phase 2}, from $t_2$ to $t_1$, both syntax and text are denoised in parallel, with the denoising models modified to enable bidirectional conditioning:
\begin{equation*}
	\begin{aligned}
		p_{\btheta_s}(\bs_{t-1}\mid \bs_{t}, \bx_t)={\mathcal N}(\bmu_{\btheta_s}(\bs_t, \bx_t,t), (1-\alpha^{s}_t)\bI),\\
		p_{\btheta_x}(\bx_{t-1}\mid \bx_{t}, \bs_t)={\mathcal N}(\bmu_{\btheta_x}(\bx_t, \bs_t,t), (1-\alpha^{x}_t)\bI).
	\end{aligned}
\end{equation*}
In \textit{Phase 3}, from $t_1$ to $0$, the model completes text denoising conditioned on the fully generated syntax via $p_{\btheta_x}(\bx_{t-1}\mid \bx_{t}, \bs_{0})$.
To maintain consistent notations across all phases, we define $\bx_{t}=\bx_{t_2}$ for $t\in[t_2, T]$ and $\bs_{t}=\bs_{t_1}$ for $t\in[0, t_1]$. The training objective for the noncascaded diffusion model is thus given by
\begin{equation}
\label{eq:noncascade_loss}
\begin{aligned}
    &{\mathcal L}_{non}(\btheta_{s},\btheta_x)
    =\mathbb{E}_{q(\bs_{0:T},\bx_{0:T})} \left[
    \log \frac{q(\bs_T \mid \bs_0)}{p_{\btheta_s}(\bs_T)}
    + \sum_{t=t_2}^{T} \log \frac{q(\bs_{t-1} \mid \bs_t, \bs_0)}{p_{\btheta_s}(\bs_{t-1} \mid \bs_t)}
    \right.\\
    &+ \sum_{t=t_1+2}^{t_2} \log \frac{q(\bs_{t-1} \mid \bs_t, \bs_0)}{p_{\btheta_s}(\bs_{t-1} \mid \bs_t, \bx_{t})}
    + \log \frac{q(\bs_0)}{p_{\btheta_s}(\bs_0 \mid \bs_{t_1+1}, \bx_{t_1+1})}
    + \log \frac{q(\bx_T \mid\bx_0)}{p_{\btheta_x}(\bx_T)}
    \\
    &\left.
    + \sum_{t=t_1+1}^{t_2} \log \frac{q(\bx_{t-1} \mid \bx_t, \bx_0)}{p_{\btheta_x}(\bx_{t-1} \mid \bx_t, \bs_{t})}
    + \sum_{t=2}^{t_1} \log \frac{q(\bx_{t-1} \mid \bx_t, \bx_0)}{p_{\btheta_x}(\bx_{t-1} \mid \bx_t, \bs_0)}
    + \log \frac{q(\bx_0)}{p_{\btheta_x}(\bx_0 \mid \bx_1, \bs_0)} \right].
\end{aligned}
\end{equation}

By allowing the generated text embeddings to influence the syntactic embeddings, the resulting structures are not only grammatically valid but also better aligned with the evolving semantic content.
In contrast to traditional cascaded approaches, this bidirectional adaptability retains the iterative refinement capability that is inherent to diffusion models, providing a more flexible and robust framework for complex generative workflows.

\begin{remark}
    By adjusting $t_1$ and $t_2$, the degree of temporal overlap can be flexibly controlled to accommodate different tasks. This allows the model to strike a balance between the mutual alignment achieved by joint refinement and the modular independence of each stage.
\end{remark}

\subsection{Complete Overlap with Unified Attention}
\label{sec:unified}
As a special case of the noncascaded framework, we consider the scenario where all stages are denoised concurrently throughout the entire trajectory by setting $t_2 = T$ and $t_1 = 0$. We refer to this configuration as \textit{complete overlap}. Under this setting, the generation of syntax and text is tightly coupled, denoising jointly at every timestep.

The complete overlap enables more efficient implementation of multi-stage generation.
In traditional diffusion, each modality requires a separate model with its own self-attention and cross-attention layers for denoising and conditioning, which increases model complexity and complicates the generation pipeline.
To improve efficiency, we propose a unified self-attention architecture that achieves mutual conditioning in a single forward pass.
Instead of preserving separate diffusion models, our architecture consolidates the denoising of syntax and text into one compact model, fully leveraging the advantage of parallel sampling.

At each timestep $t$, syntactic embeddings $\bs_t$ and text embeddings $\bx_t$ are jointly processed through a unified attention layer.
Specifically, the queries, keys, and values from the syntactic stream, $(Q(\bs_t), K(\bs_t), V(\bs_t))$, and the textual stream, $(Q(\bx_t), K(\bx_t), V(\bx_t))$, are concatenated, respectively, to form the inputs $(Q, K, V)$.
A single attention computation is then performed, producing a block-matrix attention map:
\begin{equation*}
	QK^T =
\begin{bmatrix}
Q(\bs_t)K(\bs_t)^T & Q(\bs_t)K(\bx_t)^T \\
Q(\bx_t)K(\bs_t)^T & Q(\bx_t)K(\bx_t)^T
\end{bmatrix}.
\end{equation*}
The diagonal blocks represent self-attention, capturing intra-modality interactions, while the off-diagonal blocks correspond to cross-attention, enabling inter-modality conditioning.
This unified self-attention reduces architectural redundancy and facilitates seamless bidirectional information flow between structure and content.

\begin{figure}[ht]
	\centering
	\includegraphics[width=0.42\textwidth]{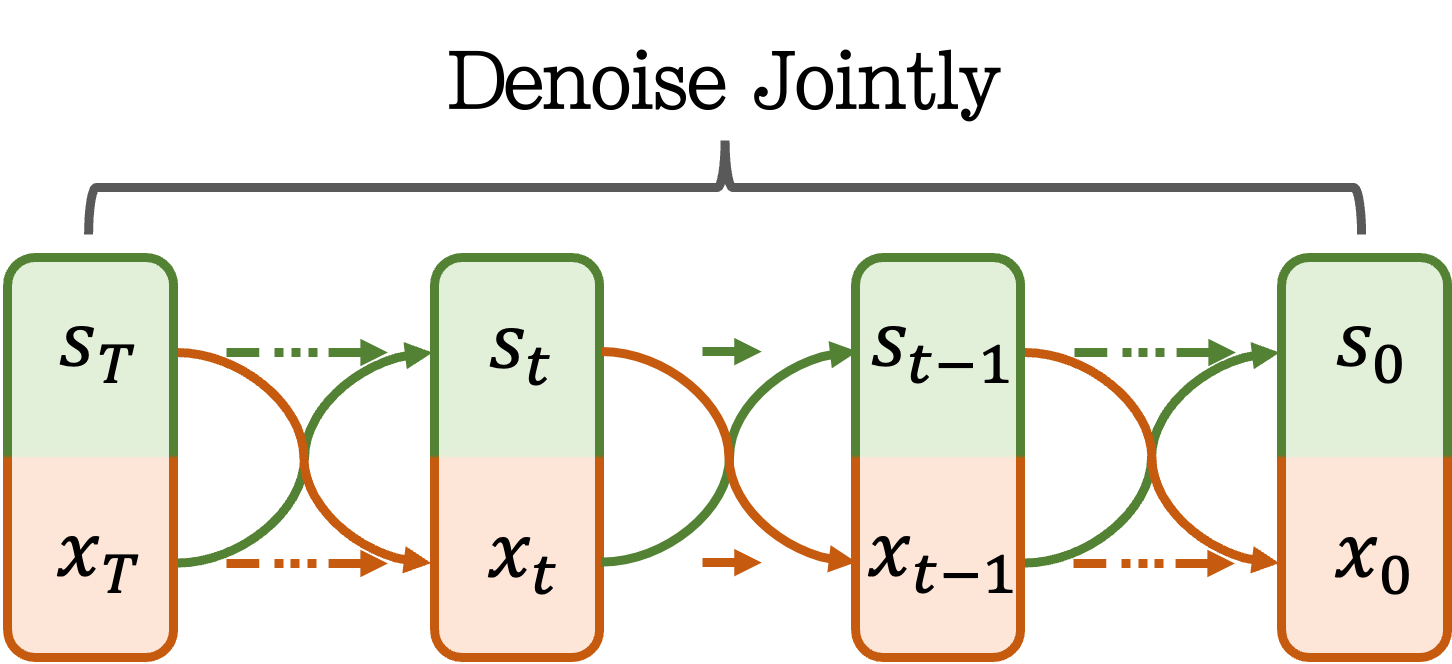}
	\caption{An illustration of the proposed \textbf{STDiff} framework. At each timestep, the syntactic and text embeddings are jointly denoised with bidirectional conditioning.}
    \label{fig:STDiff}
\end{figure}

We refer to our method as Syntax-Text Diffusion (\textbf{STDiff}), as illustrated in Figure~\ref{fig:STDiff}.
Its training objective is a special case of the general noncascaded loss in Eq.(\ref{eq:noncascade_loss}) with $t_2=T$ and $t_1=0$.
Compared to separately modeling intra- and inter-modality interactions, the unified attention layer achieves dynamic mutual conditioning while promoting parameter sharing in attention projection matrices.
For example, rather than introducing distinct syntactic key projection matrices, $W^{s}_{K,{\rm self}}$ in the syntax diffusion and $W^{s}_{K,{\rm cross}}$ in the text diffusion, we only employ a single $W^{s}_K$ within $\btheta_{s}$ to handle both roles.
Similarly, the query projection is operated through $W^{s}_Q$ in $\btheta_{s}$, replacing the need for separate $W^{s}_{Q,{\rm self}}$ and $W^{s}_{Q,{\rm cross}}$ in the syntax diffusion.

\begin{remark}
    Complete overlap expands the representational capacity of our model.
    While the cascaded framework imposes a hierarchical structure between syntax and text by modeling the true underlying joint distribution $q(\bs_0,\bx_0)$ with $p_{\btheta_x}(\bx_0 \mid \bs_0 )p_{\btheta_s}(\bs_0)$, the complete noncascaded framework relaxes this assumption by directly modeling $p_{\btheta_s, \btheta_x}(\bs_0,\bx_0)$.
    This formulation provides greater flexibility in approximating the true data distribution. We elaborate on this aspect in Section A of the Supplementary Material.
\end{remark}

\section{Experiments}
\label{sec:experiments}

We evaluate the effectiveness of our proposed cascaded and noncascaded diffusion frameworks on multiple representative real-world datasets across three aspects of text generation: quality, diversity, and personalized control.
For all experiments, we adopt the pretrained BART-base encoder-decoder \citep{lewis2020bart} for both the syntax and text branches, following the practice in \citet{Lovelace2023latent}.

\subsection{Datasets, Baselines, and Evaluation Metrics}

We conduct extensive experiments on benchmark datasets. In this section, we present the results on two representative datasets with sentiment polarity and stylistic attributes. Additional results on other datasets are provided in the Supplementary Material.

The \textbf{Yelp Review} dataset consists of user reviews labeled with sentiments, available at \url{https://business.yelp.com/}. 
We categorize reviews with 4 or 5 stars as \textit{positive}, 3 stars as \textit{neutral}, and 1 or 2 stars as \textit{negative}, thereby resulting in three distinct styles. After preprocessing, we construct a balanced training set by randomly sampling 30,000 reviews for each sentiment class.
The \textbf{Emotion} dataset \citep{saravia2018carer} contains English tweets annotated with six fine-grained emotional categories including \textit{sadness}, \textit{joy}, \textit{love}, \textit{anger}, \textit{fear}, and \textit{surprise}.
To ensure adequate syntactic richness, we filter tweets to retain those with at least 10 words, and then randomly sample 8,000 examples for each emotion.

We consider two tasks under different levels of structural conditioning: \textbf{Free Generation} and sequence-to-sequence \textbf{Sentence Expansion}. In free generation, the model synthesizes sentences directly from random noise, without explicit structural information.
In contrast, sentence expansion provides subject-verb-object (SVO) triplets as input, e.g., \textit{I love statistics}, which is more tractable with compositional guidance.

We evaluate our proposed models, \textbf{SynText} (cascaded) and \textbf{STDiff} (noncascaded), against both diffusion and autoregressive baselines.
For diffusion benchmarks, we consider \textbf{LD4LG}~\citep{Lovelace2023latent}, which generates text from compressed latent representations via auxiliary encoder-decoder modules. We also include \textbf{DiffuSeq}~\citep{gong2023diffuseq} for the sentence expansion task, which is tailored for sequence-to-sequence generation.
For autoregressive comparison, we fine-tune \textbf{GPT-2 Medium} \citep{radford2019language}, a pretrained model with a comparable parameter size, as a fair baseline of mainstream LLMs.

The evaluation metrics encompass three dimensions: generation quality, output diversity, and stylistic faithfulness. 
For generation quality, we report perplexity (\textbf{Ppl}), which reflects the negative log-likelihood of the generated text, and \textbf{Mauve} score \citep{Pillutla2021Mauve}, which quantifies the distributional similarity between the generated and reference text dataset.
We also include \textbf{BertScore} \citep{Zhang2020BERTScore} for sentence expansion, measuring semantic similarity between the generated and reference examples.
For output diversity, we compute the repetition rates of 3-grams and 4-grams (\textbf{Div-3/4}).
For stylistic faithfulness, we report the classification accuracy (\textbf{Acc}) on the generated samples measured by the DistilBERT classifier \citep{sanh2019distilbert}.
In addition, the class-wise \textbf{Mauve} scores reflect how well the generation captures the true style distribution, independent of classifier choice.
These evaluation metrics are widely used in the NLP and LLM literature \citep{bao2019syntax,mozafari2020method, Lovelace2023latent, Lou2024discrete}.

To assess the quality of syntactic learning, we introduce a new metric named syntactic n-gram overlap (\textbf{SGO}), which quantifies corpus-level n-gram similarity of syntactic patterns between generated and reference text.
Formally, we define \textbf{SGO} as the geometric mean of individual $n$-gram overlap scores $\operatorname{Overlap}(n)$ for $n = 2, 3, 4$:
\begin{equation*}
\operatorname{SGO} = \exp \left( \sum_{n\in\{2,3,4\}} \frac{1}{3} \cdot \log(\operatorname{Overlap}(n) + \epsilon) \right),
\end{equation*}
where $\epsilon$ is a small smoothing term.
Details on the definitions of $\operatorname{Overlap}(n)$ and \textbf{SGO} are provided in Section C.2 of the Supplementary Material.

\subsection{Quantitative Results}

In this subsection, we report the empirical performance of different methods on free generation and sentence expansion.
To further validate the proposed framework, Section D of the Supplementary Material provides additional experimental results, including sentence expansion on the Emotion dataset, evaluations on ROCStories and Amazon Product Reviews, ablation studies on syntactic guidance, comparisons with recent decoder-only and masked diffusion baselines, scaling analysis, computational cost comparison, and grounded generation on the iCliniq dataset.

\subsubsection{Free Generation}

In the free generation setting, models generate text from scratch solely based on a target style or label, without additional structural or lexical information.
This task allows complete flexibility in sentence composition and evaluates the model's ability to produce fluent, diverse, and stylistically appropriate text.

For evaluation, each model generates 1,000 random samples per style, which are then compared against 1,000 human-written references of the same style from an independent test set.
The results on the \textbf{Yelp Review} dataset are summarized in Table~\ref{tab:yelp_res}.

\begin{table}[ht]
\centering
\caption{Free text generation performance of different methods on the Yelp Review dataset. \textbf{Bold}
indicates the best performance, and \underline{underline} indicates the second best.}
\begin{tabular}{ccccccc}
\hline
& & \textbf{Ppl$\downarrow$} & \textbf{Mauve$\uparrow$} & \textbf{Div-3/4$\downarrow$} & \textbf{Acc$\uparrow$} &  \textbf{SGO$\uparrow$} \\  
\hline
\multirow{4}{*}{\textit{Positive}}
& SynText & \underline{81.313} & \underline{0.420} & 0.151/\underline{0.037} & \underline{0.934} & \underline{0.960} \\
& STDiff & 106.819 & \textbf{0.533} & \textbf{0.127}/\textbf{0.028} & \textbf{0.964}  & \textbf{0.962} \\
& LD4LG & 189.097 & 0.400 & \underline{0.145}/0.038 & 0.860 & 0.913 \\
& GPT-2-M & \textbf{65.727} & 0.395 & 0.242/0.132 & 0.742 & 0.845 \\
\hline
\multirow{4}{*}{\textit{Negative}}
& SynText & \underline{100.570} & 0.333 & 0.123/\underline{0.024} & \underline{0.889} & 0.942 \\
& STDiff & 131.806 & \textbf{0.421} & \textbf{0.103}/\textbf{0.018} & \textbf{0.931}  & \textbf{0.950} \\
& LD4LG & 200.224 & \underline{0.365} & \underline{0.117}/0.025 & 0.804  & 0.908 \\
& GPT-2-M & \textbf{37.498} & 0.237 & 0.267/0.139 & 0.568 & \underline{0.947} \\
\hline
\multirow{4}{*}{\textit{Neutral}}
& SynText & \underline{86.023} & \underline{0.356} & 0.155/0.036 & \underline{0.835} & \underline{0.951} \\
& STDiff & 113.021 & \textbf{0.446} & \textbf{0.138}/\textbf{0.031} & \textbf{0.894} & \textbf{0.952} \\
& LD4LG & 180.437 & 0.333 & \underline{0.143}/\underline{0.034} & 0.733  & 0.909 \\
& GPT-2-M & \textbf{49.692} & 0.162 & 0.253/0.136 & 0.560 & 0.906 \\
\hline
\end{tabular}
\label{tab:yelp_res}
\end{table}

We highlight three important observations from the results.
First, diffusion language models exhibit significantly higher diversity in generated text compared to the AR competitor, while maintaining comparable or even superior generation quality. The higher 3-gram and 4-gram repetition rates of GPT-2-M suggest its tendency to rely on frequent phrases.
Although GPT-2-M achieves lower perplexity, this metric is computed using GPT-2 and favors next-token prediction, making it less suitable for evaluating non-AR frameworks.

Second, incorporating syntax offers potential to enhance coherence and stylistic alignment. 
Compared to LD4LG, a latent diffusion based on text embeddings, our syntax-aware methods generally yield higher Mauve, SGO, and classification accuracy across all sentiment styles. 
Notably, the superior diversity of STDiff indicates that structural guidance promotes diffusion's advantage, yielding more diverse and coherent generations.

Third, the noncascaded design improves performance compared to the cascaded framework.
STDiff consistently demonstrates stronger capability in capturing both semantic and structural patterns.
While SynText shows slightly lower perplexity, this may stem from its reliance on fixed syntactic conditions, leading to more predictable word choices but less flexibility and diversity. Therefore, we adopt STDiff for subsequent comparisons.

To better evaluate stylistic learning, we assess both the ability to capture style-specific patterns and to distinguish among different styles.
We generate 1,000 samples for each sentiment and compute Mauve scores between each generated set and all reference sets to quantify their similarities.
We compare STDiff with LD4LG and GPT-2-M by reporting the pairwise relative differences in Mauve in Table~\ref{tab:yelp_mauve}, where rows indicate the sentiment of generated samples and columns indicate the reference style.
We observe that STDiff achieves higher Mauve along the diagonal, indicating its superior ability to model style-consistent distributions. Meanwhile, the off-diagonal differences are mostly negative, further suggesting that STDiff generates more distinguishable and faithful stylistic outputs.
We also observe two positive off-diagonal entries. This can be attributed to two factors. First, neutral reviews often overlap with negative ones in lexical and syntactic patterns, which naturally yields higher Mauve similarity between these two styles. Second, GPT-2-M performs poorly on the neutral class, leading to low Mauve scores against all references. 
Consequently, once STDiff captures stylistic features that are shared with negative references, the low baseline of GPT-2-M ($0.070$) amplifies the relative difference.

\begin{table}[ht]
\centering
\caption{Relative differences (\%) in Mauve scores between STDiff and comparison methods. Gen: styles of the generated text. Ref: styles of the reference text.
\textbf{Bold} indicates our method performs better than the competitor.}
\begin{tabular}{c|c}
\hline
STDiff \ vs \ LD4LG & STDiff \ vs \ GPT-2-M \\
\hline
\begin{tabular}{cccc}
Gen $\big\backslash$ Ref & \textit{Positive} & \textit{Negative} & \textit{Neutral} \\
\hline
\textit{Positive} & \textbf{33.3\%} & \textbf{-14.7\%} & \textbf{-22.0\%} \\
\textit{Negative} & \textbf{-17.9\%} & \textbf{15.7\%} & \textbf{-32.3\%} \\
\textit{Neutral} & \textbf{-31.4\%} & 4.0\% & \textbf{33.9\%} \\
\end{tabular}
&
\begin{tabular}{cccc}
Gen $\big\backslash$ Ref & \textit{Positive} & \textit{Negative} & \textit{Neutral} \\
\hline
\textit{Positive} & \textbf{34.9\%} & \textbf{-57.4\%} & \textbf{-57.4\%} \\
\textit{Negative} & \textbf{-27.3\%} & \textbf{77.6\%} & \textbf{-44.1\%} \\
\textit{Neutral} & \textbf{-7.7\%} & 84.3\% & \textbf{175.3\%} \\
\end{tabular}
\\
\hline
\end{tabular}
\label{tab:yelp_mauve}
\end{table}

The \textbf{Emotion} dataset comprises six distinct emotional styles, posing a more challenging task for faithful text personalization.
We evaluate all models under the same procedure as in the Yelp experiments, and report their performance in terms of Mauve, Acc, and SGO in Table~\ref{tab:emotion_res} to assess the quality of personalization.

\begin{table}[ht]
\centering
\caption{Free text generation performance of different methods on the Emotion dataset.}
\begin{tabular}{cccc}
\hline
 & \textit{Sadness} & \textit{Joy} & \textit{Love} \\
\hline
\begin{tabular}{c}
\\
STDiff \\
LD4LG \\
GPT-2-M \\
\end{tabular}
&
\begin{tabular}{c@{\hspace{0.8\tabcolsep}}c@{\hspace{0.8\tabcolsep}}c@{\hspace{0.8\tabcolsep}}c}
\textbf{Mauve} & \textbf{Acc} & \textbf{SGO} \\
\hline
\textbf{0.676} & \textbf{0.932}  & \textbf{0.986} \\
0.621 & 0.857 & 0.934 \\
0.495 & 0.351 & 0.895 \\
\end{tabular}
&
\begin{tabular}{c@{\hspace{0.8\tabcolsep}}c@{\hspace{0.8\tabcolsep}}c@{\hspace{0.8\tabcolsep}}c}
\textbf{Mauve} & \textbf{Acc} & \textbf{SGO}  \\
\hline
\textbf{0.748} & \textbf{0.942}  & \textbf{0.986}\\
0.533 & 0.852 & 0.921 \\
0.448 & 0.578 & 0.928 \\
\end{tabular}
&
\begin{tabular}{c@{\hspace{0.8\tabcolsep}}c@{\hspace{0.8\tabcolsep}}c}
\textbf{Mauve} & \textbf{Acc} & \textbf{SGO}  \\
\hline
\textbf{0.585} & \textbf{0.958} & \textbf{0.963} \\
0.524 & 0.899 & 0.932 \\
0.274 & 0.072 & 0.911 \\
\end{tabular}
\\
\hline
 & \textit{Anger} & \textit{Fear} & \textit{Surprise} \\
\hline
\begin{tabular}{c}
\\
STDiff \\
LD4LG \\
GPT-2-M \\
\end{tabular}
&
\begin{tabular}{c@{\hspace{0.8\tabcolsep}}c@{\hspace{0.8\tabcolsep}}c}
\textbf{Mauve} & \textbf{Acc} & \textbf{SGO} \\
\hline
\textbf{0.753} & \textbf{0.933}  & \textbf{0.986} \\
0.626 & 0.864 & 0.928 \\
0.313 & 0.852 & 0.940 \\
\end{tabular}
&
\begin{tabular}{c@{\hspace{0.8\tabcolsep}}c@{\hspace{0.8\tabcolsep}}c}
\textbf{Mauve} & \textbf{Acc} & \textbf{SGO}  \\
\hline
\textbf{0.770} & \textbf{0.942}  & \textbf{0.985}\\
0.574 & 0.821 & 0.928 \\
0.314 & 0.676 & 0.930 \\
\end{tabular}
&
\begin{tabular}{c@{\hspace{0.8\tabcolsep}}c@{\hspace{0.8\tabcolsep}}c}
\textbf{Mauve} & \textbf{Acc} & \textbf{SGO}  \\
\hline
\textbf{0.776} & \textbf{0.980} & \textbf{0.971} \\
0.538 & 0.909 & 0.909 \\
0.249 & 0.062 & 0.877 \\
\end{tabular}
\\
\hline
\end{tabular}
\label{tab:emotion_res}
\end{table}

We observe that \textbf{STDiff} consistently outperforms LD4LG and GPT-2-M across all emotional styles, exhibiting its robustness in capturing fine-grained stylistic patterns.
Compared to LD4LG, STDiff achieves significant gains in both semantic alignment and structural fidelity, suggesting that explicit syntactic guidance enhances control. In contrast, GPT-2-M struggles to identify different emotions.
These results validate the advantage of diffusion with syntax-aware personalization in handling diverse emotional identities.

\subsubsection{Sentence Expansion}
\label{sec:se}

In the sentence expansion task, the model is conditioned on structured inputs of SVO triplets.
The objective is to expand these core components into a complete sentence.
Compared to free generation, this sequence-to-sequence task offers explicit structural guidance, which reduces uncertainty in content planning.

We employ spaCy \citep{spacy2} to extract SVO triplets from human-written sentences. The extracted triplets are concatenated into a single sequence, which serves as the structural context input.
During generation, the model is conditioned on the triplets in the reference sentences to produce full sentences accordingly.
The results of all methods on the Yelp Review dataset are reported in Table~\ref{tab:yelp_res_se}.

\begin{table}[ht]
\centering
\caption{Sentence expansion performance of different methods on the Yelp Review dataset. \textbf{Bold}
indicates the best performance, and \underline{underline} indicates the second best.}
\begin{tabular}{ccccccc}
\hline
& & \textbf{BertScore$\uparrow$} & \textbf{Mauve$\uparrow$} & \textbf{Div-3/4$\downarrow$} & \textbf{Acc$\uparrow$} & \textbf{SGO$\uparrow$} \\  
\hline
\multirow{4}{*}{\textit{Positive}}
& STDiff & \textbf{0.861} & \textbf{0.904} & \textbf{0.152}/\textbf{0.038} & \textbf{0.935} & \textbf{0.986} \\
& DiffuSeq & 0.858 & 0.726 & 0.221/0.057 & 0.544 & 0.914 \\
& LD4LG & \textbf{0.861} & \underline{0.847} & \underline{0.167}/\underline{0.045} & \underline{0.913} & \underline{0.952} \\
& GPT-2-M & 0.858 & 0.397 & 0.169/0.051 & 0.833 & 0.745 \\
\hline
\multirow{4}{*}{\textit{Negative}}
& STDiff & \textbf{0.861} & \textbf{0.892} & \textbf{0.123}/\textbf{0.026} & \textbf{0.872} & \textbf{0.991} \\
& DiffuSeq & 0.858  & 0.660 & 0.197/0.044 & 0.520 & 0.904 \\
& LD4LG & \textbf{0.861} & \underline{0.754} & \underline{0.125}/\textbf{0.026} & \underline{0.843} & \underline{0.947} \\
& GPT-2-M & 0.859  & 0.483 & 0.134/0.035 & 0.787 & 0.741 \\
\hline
\multirow{4}{*}{\textit{Neutral}}
& STDiff & \underline{0.861} & \textbf{0.932} & \textbf{0.147}/\textbf{0.034} & \underline{0.827} & \textbf{0.987} \\
& DiffuSeq & 0.860 & 0.749 & 0.245/0.072 & 0.605 & 0.906 \\
& LD4LG & \textbf{0.862} & \underline{0.788} & \underline{0.170}/\underline{0.042} & \textbf{0.844}  & \underline{0.942} \\
& GPT-2-M & 0.860  & 0.475 & 0.173/0.051 & 0.712 & 0.745 \\
\hline
\end{tabular}
\label{tab:yelp_res_se}
\end{table}

We conclude that STDiff consistently achieves strong performance across all metrics and emotion categories.
Notably, it obtains the highest Mauve and SGO, along with the lowest Div-3/4, indicating its superior ability to capture both semantic and syntactic characteristics while maintaining generation diversity.
Although LD4LG performs competitively in terms of BertScore and Acc, it falls short on more comprehensive metrics, such as Mauve and SGO, as well as diversity, suggesting possible overfitting to local patterns.
Moreover, GPT-2-M exhibits significantly lower Mauve and SGO scores, highlighting the advantage of diffusion in global refinement when constructing sentences given textual fragments.
Overall, these results demonstrate the effectiveness of our method in leveraging structural inputs to improve sentence fluency and stylistic faithfulness.

\subsection{Qualitative Illustrations}

\subsubsection{Diversity and Zero-Shot Generalization}
\label{sec:qual_illu}
\begin{figure}[ht]
    \centering
    \includegraphics[width=0.9\linewidth]{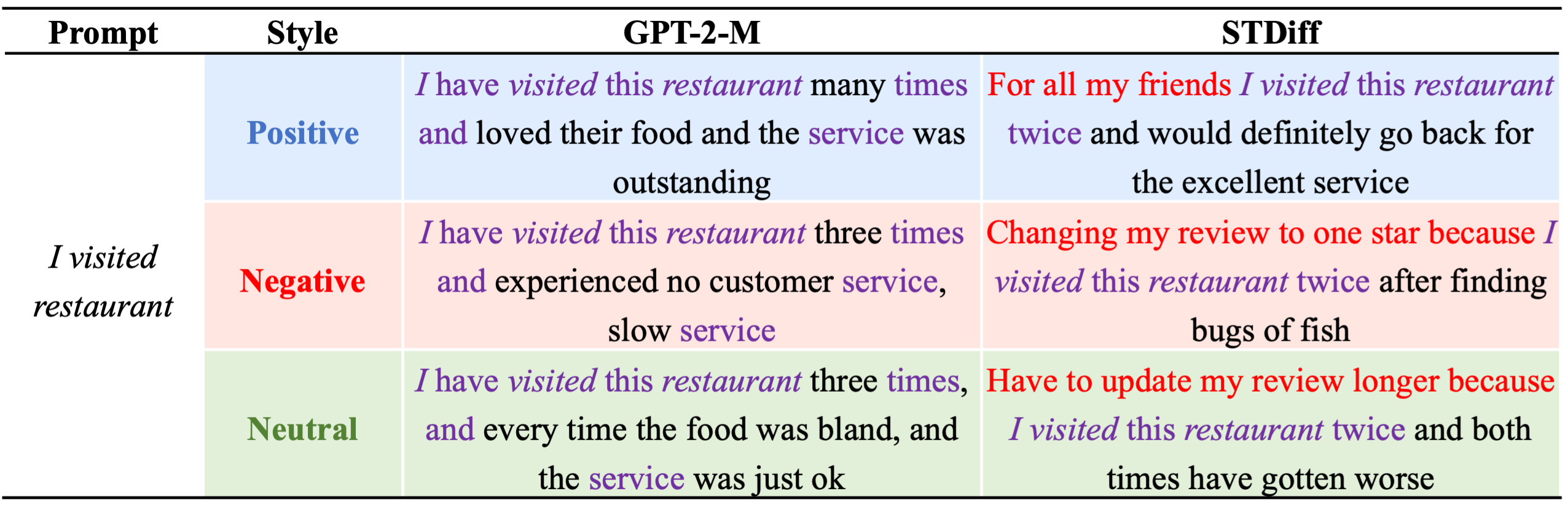}
    \caption{Sentence expansion examples from GPT-2-M and our diffusion model under the same prompt and different sentiment styles.}
    \label{fig:compare-GPT}
\end{figure}

To showcase the generating diversity and stylistic alignment of our STDiff, we present generated sentences under different sentiment styles. For better interpretability, we illustrate using the sentence expansion task, where the same SVO triplet is fed to both GPT-2-M and STDiff. The examples conditioned on sentiment labels are shown in Figure~\ref{fig:compare-GPT}.

Although GPT-2-M has a comparable number of parameters to our STDiff, its diversity and stylistic expressiveness are notably limited.
We highlight the repetitive segments in purple. The repeated fragments in STDiff outputs primarily reflect the input prompt, whereas GPT-2-M exhibits additional redundancy, including fixed sentence structures and overused content such as the word ``service.''
Moreover, the stylistic variation in GPT-2-M is mostly confined to a few sentiment-bearing words, while STDiff generates richer expressions reflecting user experience and exhibits more diverse syntactic forms.

\begin{figure}[ht]
    \centering
    \includegraphics[width=0.92\linewidth]{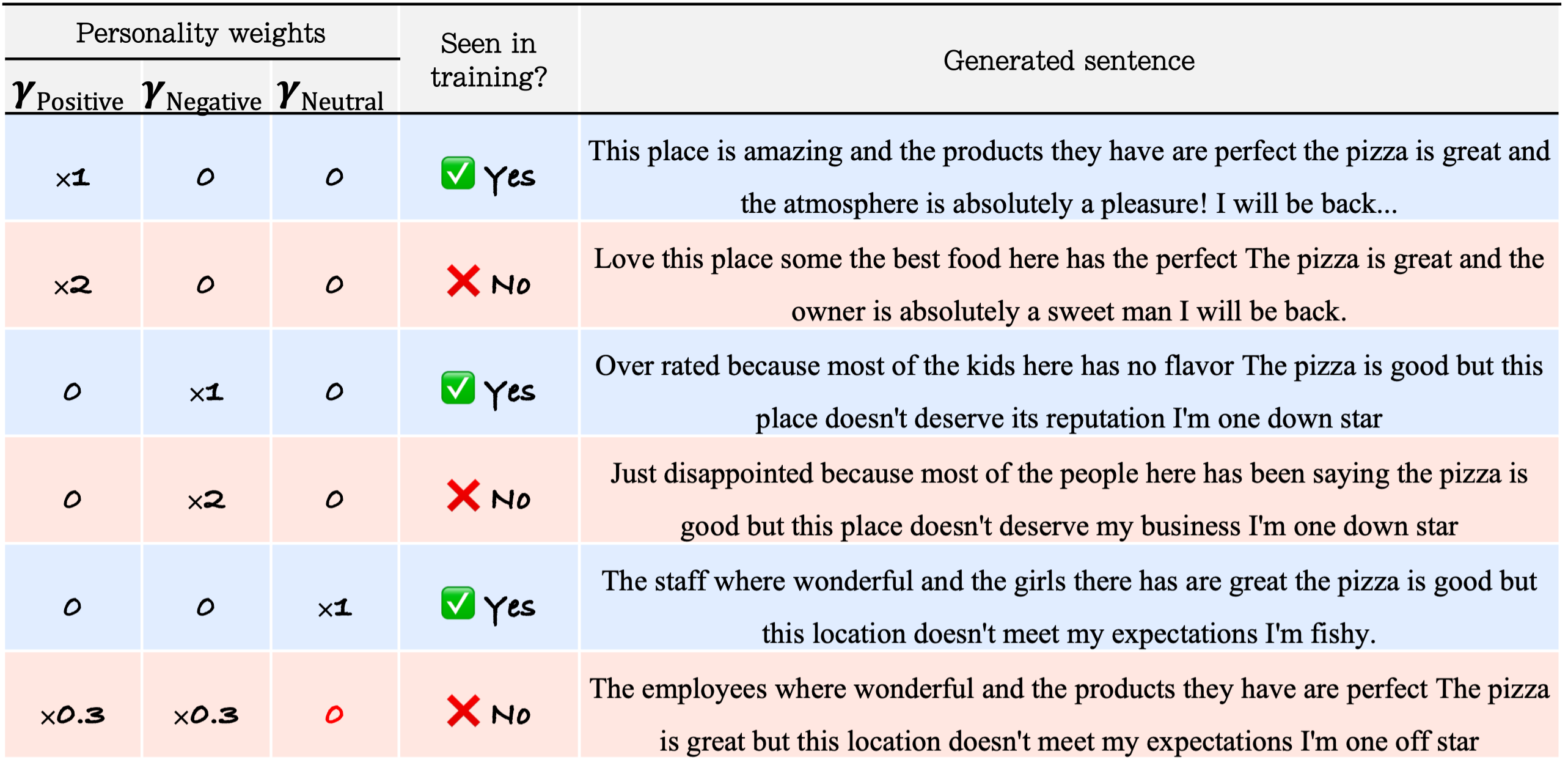}
    \caption{Zero-shot generation by adjusting personality weights. Rows highlighted in red are unseen weight combinations at training.}
    \label{fig:zero-shot}
\end{figure}

To further explore the generalizability of our personality modeling framework, we consider a zero-shot generation setting.
During training, we obtain distinct personality weights on shared representations corresponding to the canonical sentiment styles.
At the inference time, we synthesize unseen styles by combining these learned weights. 
For example, we can amplify sentiments by extrapolating, yielding stronger positive or negative styles.
We can also blend styles to create sentences reflecting intermediate characteristics and mixed tones.
The generated sentences are shown in Figure~\ref{fig:zero-shot}.
These results demonstrate that our sharing framework learns a unified latent space of personalities, enabling smooth generalization beyond training styles, which contrasts with conventional approaches that learn discrete embeddings for each style independently.

\subsubsection{Personalization Visualization}

\begin{figure}[ht]
    \centering
    \begin{subfigure}[b]{0.49\linewidth}
        \centering
        \includegraphics[width=\linewidth]{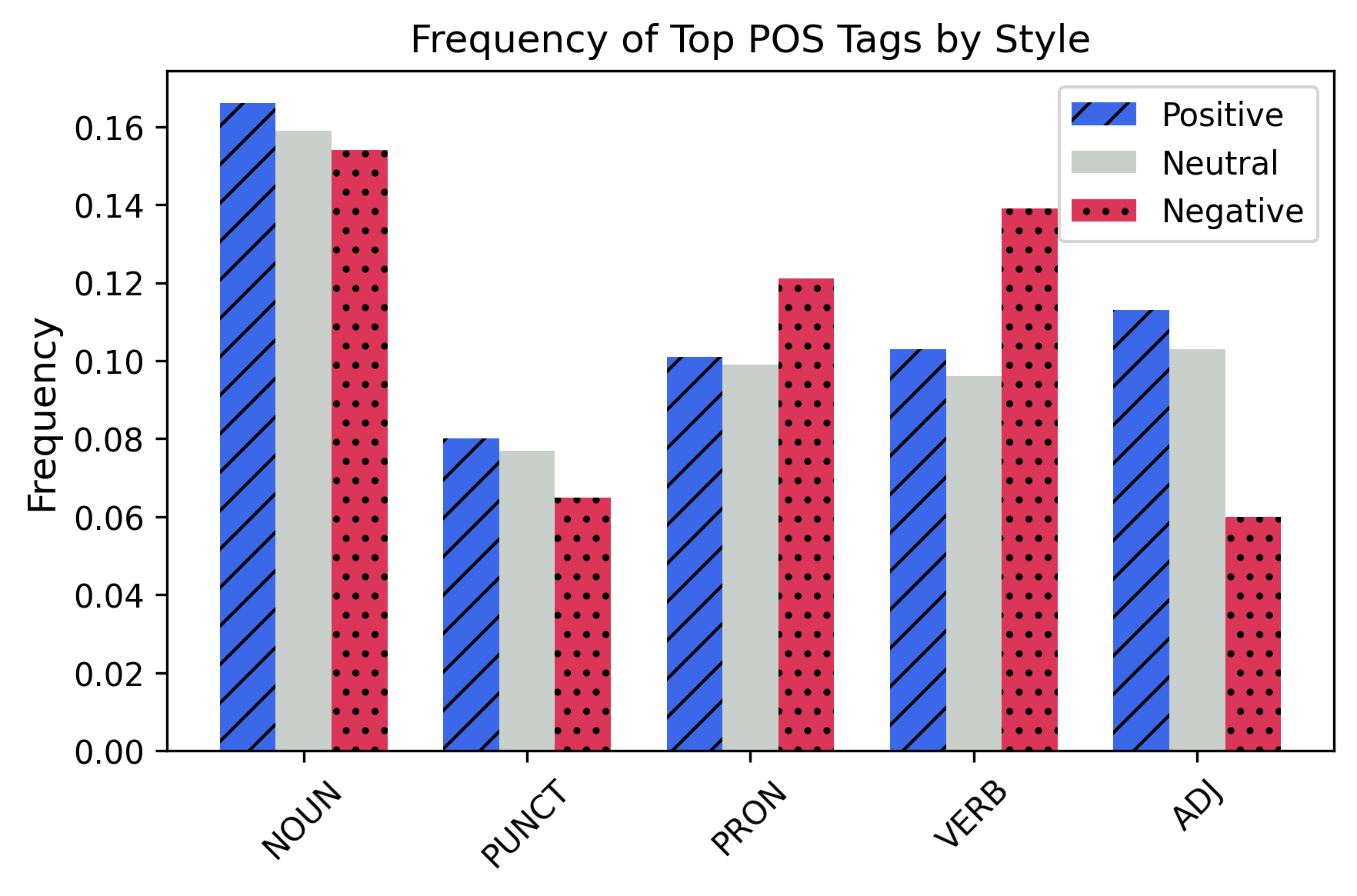}
    \end{subfigure}
    \hfill
    \begin{subfigure}[b]{0.49\linewidth}
        \centering
        \includegraphics[width=0.97\linewidth]{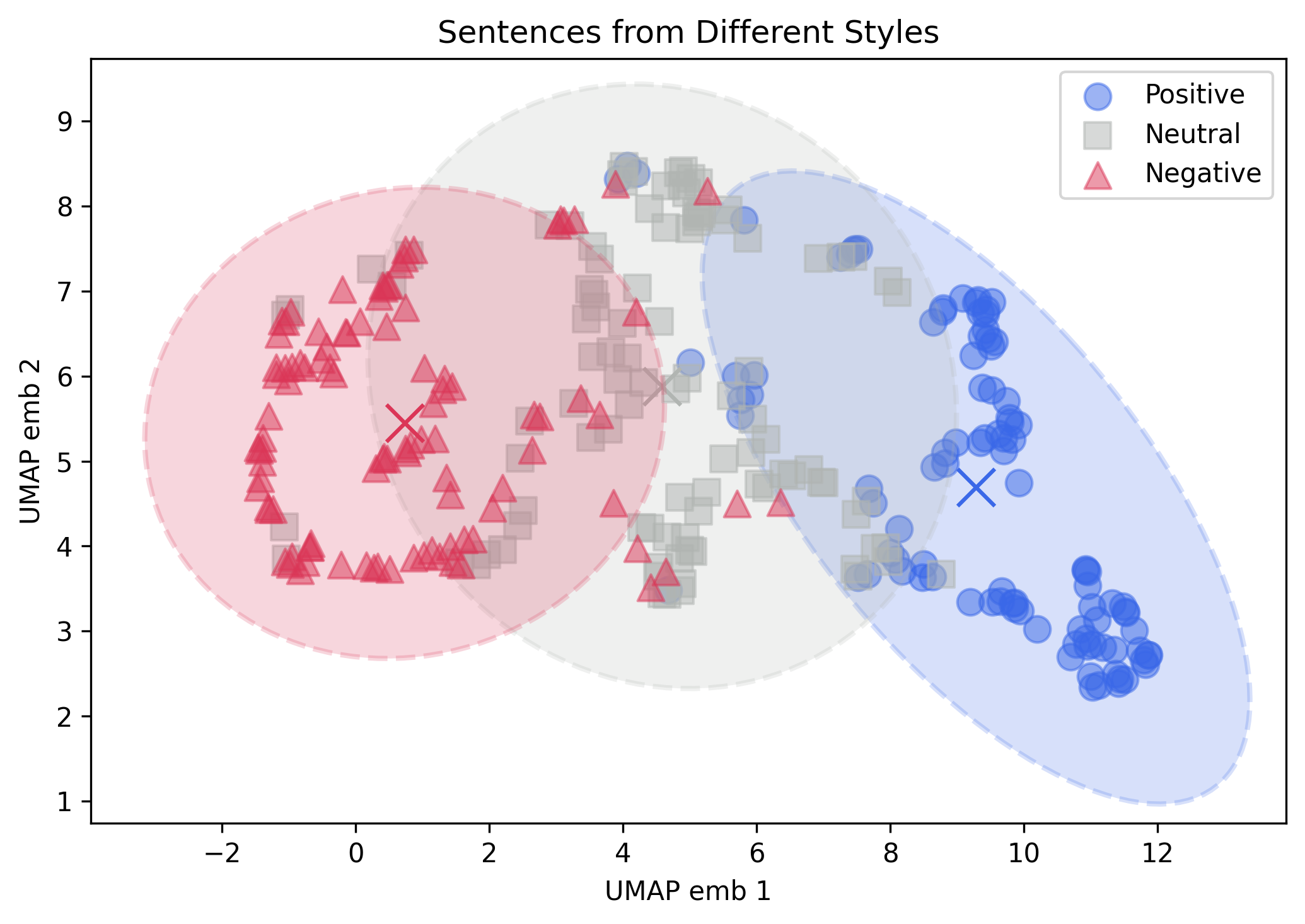}
    \end{subfigure}
    \caption{Frequencies of POS tags and UMAP of sentence embeddings in generated text.}
    \label{fig:gen-pattern}
\end{figure}

To illustrate the effectiveness of our method in capturing human-written patterns, we analyze the generated text from both structural and semantic perspectives on the Yelp Review dataset.
In Figure~\ref{fig:gen-pattern}, we examine syntactic structures by measuring POS frequencies for each style, and visualize sentence embeddings using outputs from the text diffusion.

We observe that the generated text closely resembles the reference data in the motivating examples (see Figure~\ref{fig:yelp-pos-freq} and \ref{fig:yelp-text}), demonstrating the ability to preserve underlying distributions of each style.
Moreover, compared to real data, our method exhibits clearer separation between different styles, particularly in the embedding space. This suggests that our model enhances inter-style distinctiveness, resulting in more style-consistent generation.

\begin{figure}[ht]
    \centering
    \includegraphics[width=0.7\linewidth]{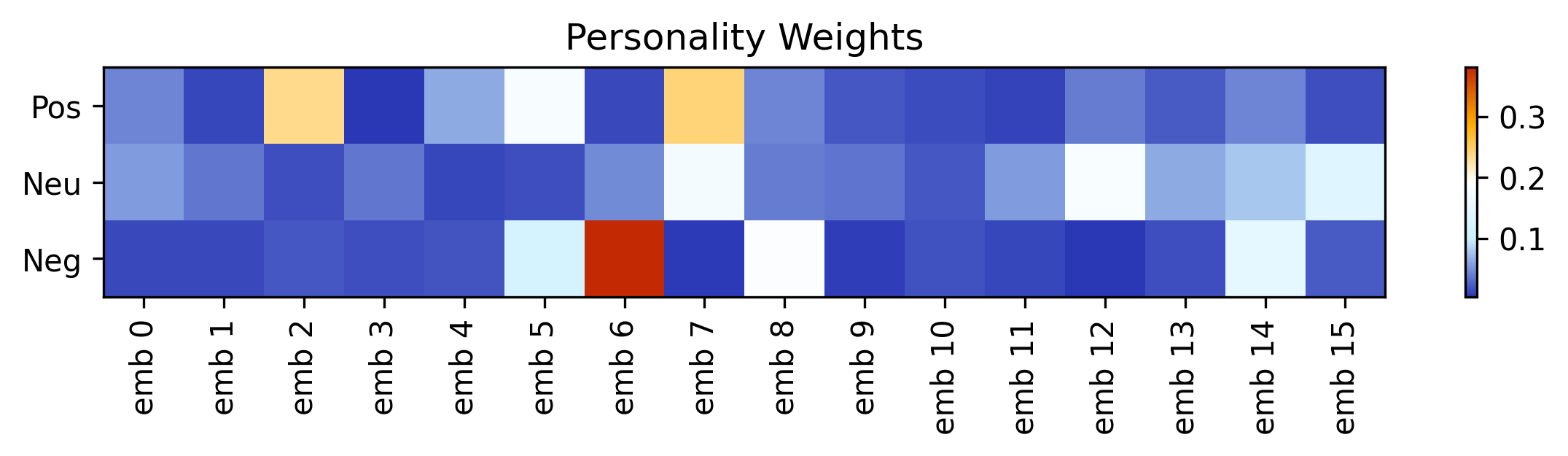}
    \caption{Personality weights for different styles over different personality embeddings.}
    \label{fig:weights}
\end{figure}

To better interpret the learned personality representations, we visualize the personality weights for each sentiment style in Figure~\ref{fig:weights}. While all styles share certain common embeddings, each style is characterized by distinctive key components. For instance, embeddings 2 and 7 are more prominent in the \textit{Positive} style, whereas embedding 6 is more active in the \textit{Negative} style. Furthermore, the \textit{Neutral} weights exhibit an intermediate pattern, indicating the model's ability to learn a semantically meaningful latent space.

\section{Discussion}
\label{sec:discuss}

The growing concern over linguistic homogenization in LLMs underscores the need for mechanisms which restore structural diversity and personalization in text generation.
In this study, we explore how statistical modeling principles can enhance large language models in a fundamental and interpretable manner.
By modeling the text distribution hierarchically, structural information is incorporated into the generative process, mirroring the natural formation of syntax and semantics.
Through latent representation learning, the proposed method pools information across styles while achieving a personalized language model with improved data efficiency and interpretability.
Overall, from a statistical perspective, the proposed framework can be viewed as a structured latent-variable decomposition of text generation that separates global syntactic organization from local lexical realization, while enabling shared representation learning across related stylistic domains.
Experiments on real-world text corpora confirm the advantages of our proposed framework under comparable parameter scales, suggesting a promising direction for structure-aware text generation.

The syntactic guidance proposed in this work provides a general principle for incorporating structural information into text generation. 
Although our current implementation adopts a continuous diffusion formulation, the core ideas of structural conditioning and shared personalized representations are highly generalizable and may be extended to broader language model families. 
One promising direction is to incorporate syntactic information into discrete diffusion language models, which have recently attracted growing attention \citep{MDLM, Lou2024discrete, yu2025discrete}.
The relatively small vocabulary size of syntactic entities may further enable efficient and stable discrete sampling.
In addition, while we employ POS tags as a reliable form of syntactic supervision, future studies could utilize more fine-grained signals, such as dependency relations and constituency trees.

Another future direction lies in refining the noncascaded architecture.
As a generalization of the cascaded framework, the noncascaded design is applicable beyond our current setting and can extend to a broad range of hierarchical generation scenarios, including multi-stage and multi-modal generation.
Different applications may benefit from varying degrees of overlap between stages. A systematic exploration of information flow across different stages can provide deeper insights into the trade-offs between sequential and parallel generation, thereby motivating new large language model frameworks.

Several limitations should also be acknowledged. First, although the proposed framework improves syntactic diversity and personalized control, diffusion-based language models remain computationally more expensive than autoregressive models because of iterative denoising during generation. Second, while the proposed models demonstrate strong performance at moderate parameter scales, experiments were not conducted at the scale of frontier commercial LLMs, and the behavior of syntax-guided diffusion under substantially larger architectures remains an important direction for future work. Third, the current framework relies on part-of-speech representations extracted from external parsers, and syntactic guidance quality may therefore depend on parser accuracy. Finally, most experiments focus on sentence-level or moderately long-form generation tasks, and additional evaluation on broader reasoning-intensive or multi-turn generation settings would further clarify the strengths and limitations of syntax-guided diffusion language modeling.

\section*{Significance Statement}
Modern language models often produce fluent but structurally repetitive text and offer limited control over individual writing styles. We develop a text-generation framework that first models sentence structure and then generates words consistent with that structure. Style information enters both stages, providing control over sentence form and word choice, while shared style components allow related styles to borrow information from one another. Unlike existing personalization methods that mainly adjust word choice or learn each style separately, our approach produces more diverse and style-consistent text and can generate new styles by combining learned patterns. More broadly, it offers a general way to integrate structural guidance and personalization in generative models.

\begin{acks}[Acknowledgments]
The authors would like to thank the referees, an Associate Editor and the Editor for their constructive comments that improved the quality of this paper.
\end{acks}

\begin{funding}
{
The research is supported by NSF Grant CDS\&E-MSS 2401271.
}
\end{funding}

\begin{supplement}
\stitle{Supplementary analyses and methodological details} \sdescription{The Supplementary Material includes a discussion on the improvement from the noncascaded framework, implementation details, additional information on syntax and metrics, additional experiments, and representative generated examples.}
\end{supplement}

\begin{supplement}
\stitle{Code repository} \sdescription{Python files
used to generate the results of this paper, which is publicly available at \url{https://github.com/RuqianZhang/syntax-diffusion}.}
\end{supplement}


\bibliographystyle{imsart-nameyear} 
\bibliography{main}       

@misc{xu2025representation,
      title={Representation Retrieval Learning for Heterogeneous Data Integration}, 
      author={Qi Xu and Annie Qu},
      year={2025},
      eprint={2503.09494},
      archivePrefix={arXiv},
      primaryClass={cs.LG},
      url={https://arxiv.org/abs/2503.09494}, 
}

@article{xie2020represent,
   author = "Xie, Jianwen and Gao, Ruiqi and Nijkamp, Erik and Zhu, Song-Chun and Wu, Ying Nian",
   title = "Representation Learning: A Statistical Perspective", 
   journal= "Annual Review of Statistics and Its Application",
   year = "2020",
   volume = "7",
   number = "Volume 7, 2020",
   pages = "303-335"
  }

@inproceedings{Bachman2016,
 author = {Bachman, Philip},
 booktitle = {Advances in Neural Information Processing Systems},
 pages = {},
 title = {An Architecture for Deep, Hierarchical Generative Models},
 volume = {29},
 year = {2016}
}

@inproceedings{mozafari2020method,
  title={A method for answer selection using DistilBERT and important words},
  author={Mozafari, Jamshid and Fatemi, Afsaneh and Moradi, Parham},
  booktitle={2020 6th International Conference on Web Research (ICWR)},
  pages={72--76},
  year={2020},
  organization={IEEE}
}

@article {Mummery368,
	author = {Mummery, David},
	title = {Artificial intelligence {\textellipsis} or not?},
	volume = {75},
	number = {757},
	pages = {368--368},
	year = {2025},
	journal = {British Journal of General Practice}
}

@misc{labs2025mercury,
      title={Mercury: Ultra-Fast Language Models Based on Diffusion}, 
      author={InceptionLabs and Samar Khanna and Siddhant Kharbanda and Shufan Li and Harshit Varma and Eric Wang and et al.},
      year={2025},
      eprint={2506.17298},
      archivePrefix={arXiv},
      primaryClass={cs.CL},
      url={https://arxiv.org/abs/2506.17298}, 
}

@misc{deepseekai2025deepseekv3,
      title={DeepSeek-V3 Technical Report}, 
      author={DeepSeek-AI and Aixin Liu and Bei Feng and Bing Xue and Bingxuan Wang and Bochao Wu and et al.},
      year={2025},
      eprint={2412.19437},
      archivePrefix={arXiv},
      primaryClass={cs.CL},
      url={https://arxiv.org/abs/2412.19437}, 
}

@misc{openai2024gpt4ocard,
      title={GPT-4o System Card}, 
      author={OpenAI and Aaron Hurst and Adam Lerer and Adam P. Goucher and Adam Perelman and Aditya Ramesh and et al.},
      year={2024},
      eprint={2410.21276},
      archivePrefix={arXiv},
      primaryClass={cs.CL},
      url={https://arxiv.org/abs/2410.21276}, 
}

@inproceedings{wan2023kelly,
    title = "``{Kelly} is a Warm Person, {Joseph} is a Role Model'': Gender Biases in {LLM}-Generated Reference Letters",
    author = "Wan, Yixin  and
      Pu, George  and
      Sun, Jiao  and
      Garimella, Aparna  and
      Chang, Kai-Wei  and
      Peng, Nanyun",
    booktitle = "Findings of the Association for Computational Linguistics: EMNLP 2023",
    year = "2023",
    pages = "3730--3748",
}

@inproceedings{Lou2024discrete,
author = {Lou, Aaron and Meng, Chenlin and Ermon, Stefano},
title = {Discrete diffusion modeling by estimating the ratios of the data distribution},
year = {2024},
booktitle = {Proceedings of the 41st International Conference on Machine Learning},
articleno = {1333},
numpages = {30}
}

@misc{yu2025discrete,
      title={Discrete Diffusion in Large Language and Multimodal Models: A Survey}, 
      author={Runpeng Yu and Qi Li and Xinchao Wang},
      year={2025},
      eprint={2506.13759},
      archivePrefix={arXiv},
      primaryClass={cs.LG},
      url={https://arxiv.org/abs/2506.13759}, 
}

@INPROCEEDINGS{hong2018inferring,
  author={Hong, Seunghoon and Yang, Dingdong and Choi, Jongwook and Lee, Honglak},
  booktitle={2018 IEEE/CVF Conference on Computer Vision and Pattern Recognition}, 
  title={Inferring Semantic Layout for Hierarchical Text-to-Image Synthesis}, 
  year={2018},
  pages={7986-7994}
}

@InProceedings{hayou24lora,
  title = 	 {{L}o{RA}+: Efficient Low Rank Adaptation of Large Models},
  author =       {Hayou, Soufiane and Ghosh, Nikhil and Yu, Bin},
  booktitle = 	 {Proceedings of the 41st International Conference on Machine Learning},
  pages = 	 {17783--17806},
  year = 	 {2024},
  volume = 	 {235},
}

@misc{mcinnes2020umap,
      title={UMAP: Uniform Manifold Approximation and Projection for Dimension Reduction}, 
      author={Leland McInnes and John Healy and James Melville},
      year={2020},
      eprint={1802.03426},
      archivePrefix={arXiv},
      primaryClass={stat.ML},
      url={https://arxiv.org/abs/1802.03426}, 
}

@inproceedings{gao2021simcse,
    title = "{S}im{CSE}: Simple Contrastive Learning of Sentence Embeddings",
    author = "Gao, Tianyu  and
      Yao, Xingcheng  and
      Chen, Danqi",
    booktitle = "Proceedings of the 2021 Conference on Empirical Methods in Natural Language Processing",
    year = "2021",
    pages = "6894--6910"
}

@InProceedings{frenkel2024implicit,
author="Frenkel, Yarden
and Vinker, Yael
and Shamir, Ariel
and Cohen-Or, Daniel",
title="Implicit Style-Content Separation Using B-LoRA",
booktitle="European Conference on Computer Vision",
year="2025",
publisher="Springer",
pages="181--198",
}

@inproceedings{alhafni2024personalized,
    title = "Personalized Text Generation with Fine-Grained Linguistic Control",
    author = "Alhafni, Bashar  and
      Kulkarni, Vivek  and
      Kumar, Dhruv  and
      Raheja, Vipul",
    booktitle = "Proceedings of the 1st Workshop on Personalization of Generative AI Systems (PERSONALIZE 2024)",
    year = "2024",
    url = "https://aclanthology.org/2024.personalize-1.8/",
    pages = "88--101"
}

@book{Chomsky1957,
title = {Syntactic Structures},
author = {Noam Chomsky},
publisher = {De Gruyter Mouton},
address = {Berlin, Boston},
year = {1957}
}

@inproceedings{bao2019syntax,
    title = "Generating Sentences from Disentangled Syntactic and Semantic Spaces",
    author = "Bao, Yu  and
      Zhou, Hao  and
      Huang, Shujian  and
      Li, Lei  and
      Mou, Lili  and
      Vechtomova, Olga  and
      et al.",
    booktitle = "Proceedings of the 57th Annual Meeting of the Association for Computational Linguistics",
    year = "2019",
    pages = "6008--6019"
}

@inproceedings{li2023Syntactic,
    title = "Explicit Syntactic Guidance for Neural Text Generation",
    author = "Li, Yafu  and
      Cui, Leyang  and
      Yan, Jianhao  and
      Yin, Yongjing  and
      Bi, Wei  and
      Shi, Shuming  and
      Zhang, Yue",
    booktitle = "Proceedings of the 61st Annual Meeting of the Association for Computational Linguistics",
    year = "2023",
    pages = "14095--14112"
}

@misc{rout2025anchored,
      title={Anchored Diffusion Language Model}, 
      author={Litu Rout and Constantine Caramanis and Sanjay Shakkottai},
      year={2025},
      eprint={2505.18456},
      archivePrefix={arXiv},
      primaryClass={cs.CL},
      url={https://arxiv.org/abs/2505.18456}, 
}

@inproceedings{Holtzman2020curious,
  title={The Curious Case of Neural Text Degeneration},
  author={Holtzman, Ari and Buys, Jan and Du, Li and Forbes, Maxwell and Choi, Yejin},
  year={2020},
  booktitle={International Conference on Learning Representations}
}

@misc{sourati2025diversity,
      title={The Shrinking Landscape of Linguistic Diversity in the Age of Large Language Models}, 
      author={Zhivar Sourati and Farzan Karimi-Malekabadi and Meltem Ozcan and Colin McDaniel and Alireza Ziabari and Jackson Trager and et al},
      year={2025},
      eprint={2502.11266},
      archivePrefix={arXiv},
      primaryClass={cs.CL},
      url={https://arxiv.org/abs/2502.11266}, 
}

@misc{nie2025large,
      title={Large Language Diffusion Models}, 
      author={Shen Nie and Fengqi Zhu and Zebin You and Xiaolu Zhang and Jingyang Ou and Jun Hu and et al},
      year={2025},
      eprint={2502.09992},
      archivePrefix={arXiv},
      primaryClass={cs.CL},
      url={https://arxiv.org/abs/2502.09992}, 
}

@inproceedings{Saharia2022imagen,
 author = {Saharia, Chitwan and Chan, William and Saxena, Saurabh and Li, Lala and Whang, Jay and Denton, Emily L and et al},
 booktitle = {Advances in Neural Information Processing Systems},
 pages = {36479--36494},
 title = {Photorealistic Text-to-Image Diffusion Models with Deep Language Understanding},
 volume = {35},
 year = {2022}
}

@inproceedings{Song2019score,
 author = {Song, Yang and Ermon, Stefano},
 booktitle = {Advances in Neural Information Processing Systems},
 pages = {},
 title = {Generative Modeling by Estimating Gradients of the Data Distribution},
 volume = {32},
 year = {2019}
}

@inproceedings{Li2023survey,
  title     = {Diffusion Models for Non-autoregressive Text Generation: A Survey},
  author    = {Li, Yifan and Zhou, Kun and Zhao, Wayne Xin and Wen, Ji-Rong},
  booktitle = {Proceedings of the Thirty-Second International Joint Conference on
               Artificial Intelligence, {IJCAI-23}},
  pages     = {6692--6701},
  year      = {2023},
  note      = {Survey Track}
}

@article{blei2003latent,
  title={Latent dirichlet allocation},
  author={Blei, David M and Ng, Andrew Y and Jordan, Michael I},
  journal={Journal of machine Learning research},
  volume={3},
  number={Jan},
  pages={993--1022},
  year={2003}
}

@article{radford2019language,
  title={Language models are unsupervised multitask learners},
  author={Radford, Alec and Wu, Jeffrey and Child, Rewon and Luan, David and Amodei, Dario and Sutskever, Ilya and others},
  journal={OpenAI blog},
  volume={1},
  number={8},
  pages={9},
  year={2019}
}

@inproceedings{saravia2018carer,
    title = "{CARER}: Contextualized Affect Representations for Emotion Recognition",
    author = "Saravia, Elvis  and
      Liu, Hsien-Chi Toby  and
      Huang, Yen-Hao  and
      Wu, Junlin  and
      Chen, Yi-Shin",
    booktitle = "Proceedings of the 2018 Conference on Empirical Methods in Natural Language Processing",
    year = "2018",
    pages = "3687--3697"
}

@misc{ramesh2022unclip,
      title={Hierarchical Text-Conditional Image Generation with CLIP Latents}, 
      author={Aditya Ramesh and Prafulla Dhariwal and Alex Nichol and Casey Chu and Mark Chen},
      year={2022},
      eprint={2204.06125},
      archivePrefix={arXiv},
      primaryClass={cs.CV},
      url={https://arxiv.org/abs/2204.06125}, 
}

@inproceedings{Ho2020DDPM,
 author = {Ho, Jonathan and Jain, Ajay and Abbeel, Pieter},
 booktitle = {Advances in Neural Information Processing Systems},
 pages = {6840--6851},
 title = {Denoising Diffusion Probabilistic Models},
 volume = {33},
 year = {2020}
}

@inproceedings{Hoogeboom2021,
 author = {Hoogeboom, Emiel and Nielsen, Didrik and Jaini, Priyank and Forr\'{e}, Patrick and Welling, Max},
 booktitle = {Advances in Neural Information Processing Systems},
 pages = {12454--12465},
 title = {Argmax Flows and Multinomial Diffusion: Learning Categorical Distributions},
 volume = {34},
 year = {2021}
}

@inproceedings{chen2023analog,
title={Analog Bits: Generating Discrete Data using Diffusion Models with Self-Conditioning},
author={Ting Chen and Ruixiang Zhang and Geoffrey Hinton},
booktitle={The Eleventh International Conference on Learning Representations},
year={2023}
}

@inproceedings{Li2022LM,
 author = {Li, Xiang and Thickstun, John and Gulrajani, Ishaan and Liang, Percy S and Hashimoto, Tatsunori B},
 booktitle = {Advances in Neural Information Processing Systems},
 pages = {4328--4343},
 title = {Diffusion-{LM} Improves Controllable Text Generation},
 volume = {35},
 year = {2022}
}

@inproceedings{gong2023diffuseq,
title={DiffuSeq: Sequence to Sequence Text Generation with Diffusion Models},
author={Shansan Gong and Mukai Li and Jiangtao Feng and Zhiyong Wu and Lingpeng Kong},
booktitle={The Eleventh International Conference on Learning Representations},
year={2023}
}

@inproceedings{Lovelace2023latent,
 author = {Lovelace, Justin and Kishore, Varsha and Wan, Chao and Shekhtman, Eliot and Weinberger, Kilian Q},
 booktitle = {Advances in Neural Information Processing Systems},
 pages = {56998--57025},
 title = {Latent Diffusion for Language Generation},
 volume = {36},
 year = {2023}
}

@InProceedings{Peebles2023DiT,
    author    = {Peebles, William and Xie, Saining},
    title     = {Scalable Diffusion Models with Transformers},
    booktitle = {Proceedings of the IEEE/CVF International Conference on Computer Vision (ICCV)},
    year      = {2023},
    pages     = {4195-4205}
}

@article{Ho2022cascade,
  author  = {Jonathan Ho and Chitwan Saharia and William Chan and David J. Fleet and Mohammad Norouzi and Tim Salimans},
  title   = {Cascaded Diffusion Models for High Fidelity Image Generation},
  journal = {Journal of Machine Learning Research},
  year    = {2022},
  volume  = {23},
  number  = {47},
  pages   = {1--33},
  url     = {http://jmlr.org/papers/v23/21-0635.html}
}

@inproceedings{salemi2024lamp,
    title = "{L}a{MP}: When Large Language Models Meet Personalization",
    author = "Salemi, Alireza  and
      Mysore, Sheshera  and
      Bendersky, Michael  and
      Zamani, Hamed",
    booktitle = "Proceedings of the 62nd Annual Meeting of the Association for Computational Linguistics",
    year = "2024",
    pages = "7370--7392"
}

@inproceedings{zhong2021useradapter,
    title = "{U}ser{A}dapter: Few-Shot User Learning in Sentiment Analysis",
    author = "Zhong, Wanjun  and
      Tang, Duyu  and
      Wang, Jiahai  and
      Yin, Jian  and
      Duan, Nan",
    booktitle = "Findings of the Association for Computational Linguistics: ACL-IJCNLP 2021",
    year = "2021",
    pages = "1484--1488"
}

@inproceedings{mireshghallah2022useridentifier,
    title = "{U}ser{I}dentifier: Implicit User Representations for Simple and Effective Personalized Sentiment Analysis",
    author = "Mireshghallah, Fatemehsadat  and
      Shrivastava, Vaishnavi  and
      Shokouhi, Milad  and
      Berg-Kirkpatrick, Taylor  and
      Sim, Robert  and
      Dimitriadis, Dimitrios",
    booktitle = "Proceedings of the 2022 Conference of the North American Chapter of the Association for Computational Linguistics",
    year = "2022",
    pages = "3449--3456"
}

@inproceedings{hu2022lora,
title={Lo{RA}: Low-Rank Adaptation of Large Language Models},
author={Edward J Hu and Yelong Shen and Phillip Wallis and Zeyuan Allen-Zhu and Yuanzhi Li and Shean Wang and Lu Wang and Weizhu Chen},
booktitle={International Conference on Learning Representations},
year={2022}
}

@unpublished{spacy2,
    AUTHOR = {Honnibal, Matthew and Montani, Ines},
    TITLE  = {{spaCy 2}: Natural language understanding with {B}loom embeddings, convolutional neural networks and incremental parsing},
    YEAR   = {2017},
    Note   = {To appear}
}

@inproceedings{Zhang2020BERTScore,
title={BERTScore: Evaluating Text Generation with BERT},
author={Tianyi Zhang and Varsha Kishore and Felix Wu and Kilian Q. Weinberger and Yoav Artzi},
booktitle={International Conference on Learning Representations},
year={2020}
}

@inproceedings{Pillutla2021Mauve,
 author = {Pillutla, Krishna and Swayamdipta, Swabha and Zellers, Rowan and Thickstun, John and Welleck, Sean and Choi, Yejin and Harchaoui, Zaid},
 booktitle = {Advances in Neural Information Processing Systems},
 pages = {4816--4828},
 title = {MAUVE: Measuring the Gap Between Neural Text and Human Text using Divergence Frontiers},
 volume = {34},
 year = {2021}
}

@inproceedings{sanh2019distilbert,
  title={DistilBERT, a distilled version of BERT: smaller, faster, cheaper and lighter},
  booktitle = {Advances in Neural Information Processing Systems Workshop on Energy Efficient Machine Learning and Cognitive Computing},
  author={Victor Sanh and Lysandre Debut and Julien Chaumond and Thomas Wolf},
  year={2019}
}

@inproceedings{Vaswani2017,
 author = {Vaswani, Ashish and Shazeer, Noam and Parmar, Niki and Uszkoreit, Jakob and Jones, Llion and Gomez, Aidan N and Kaiser, \L ukasz and Polosukhin, Illia},
 booktitle = {Advances in Neural Information Processing Systems},
 pages = {},
 title = {Attention is All you Need},
 volume = {30},
 year = {2017}
}

@inproceedings{lewis2020bart,
    title = "{BART}: Denoising Sequence-to-Sequence Pre-training for Natural Language Generation, Translation, and Comprehension",
    author = "Lewis, Mike  and
      Liu, Yinhan  and
      Goyal, Naman  and
      Ghazvininejad, Marjan  and
      Mohamed, Abdelrahman  and
      Levy, Omer  and
      Stoyanov, Veselin  and
      Zettlemoyer, Luke",
    booktitle = "Proceedings of the 58th Annual Meeting of the Association for Computational Linguistics",
    year = "2020",
    pages = "7871--7880"
}

@inproceedings{MDLM,
 author = {Shi, Jiaxin and Han, Kehang and Wang, Zhe and Doucet, Arnaud and Titsias, Michalis},
 booktitle = {Advances in Neural Information Processing Systems},
 pages = {103131--103167},
 publisher = {Curran Associates, Inc.},
 title = {Simplified and Generalized Masked Diffusion for Discrete Data},
 volume = {37},
 year = {2024}
}

@article{Fan2013lowrank,
    author = {Fan, Jianqing and Liao, Yuan and Mincheva, Martina},
    title = {Large Covariance Estimation by Thresholding Principal Orthogonal Complements},
    journal = {Journal of the Royal Statistical Society Series B: Statistical Methodology},
    volume = {75},
    number = {4},
    pages = {603-680},
    year = {2013}
}

@article{Chen01122012,
author = {Lisha Chen and Jianhua Z. Huang},
title = {Sparse Reduced-Rank Regression for Simultaneous Dimension Reduction and Variable Selection},
journal = {Journal of the American Statistical Association},
volume = {107},
number = {500},
pages = {1533--1545},
year = {2012},
publisher = {Taylor \& Francis}
}

@article{Blei2014review,
   author = "Blei, David M.",
   title = "Build, Compute, Critique, Repeat: Data Analysis with Latent Variable Models", 
   journal= "Annual Review of Statistics and Its Application",
   year = "2014",
   volume = "1",
   number = "Volume 1, 2014",
   pages = "203-232",
   publisher = "Annual Reviews",
   type = "Journal Article"
  }

@article{Noirrit2023latent,
  author  = {Noirrit Kiran Chandra and Antonio Canale and David B. Dunson},
  title   = {Escaping The Curse of Dimensionality in {B}ayesian Model-Based Clustering},
  journal = {Journal of Machine Learning Research},
  year    = {2023},
  volume  = {24},
  number  = {144},
  pages   = {1--42},
  url     = {http://jmlr.org/papers/v24/21-1276.html}
}

@article{Sui2025multi,
author = {Yang Sui and Qi Xu and Yang Bai and Annie Qu},
title = {Multi-Task Learning for Heterogeneous Multi-Source Block-Wise Missing Data},
journal = {Journal of Computational and Graphical Statistics},
volume = {0},
number = {0},
pages = {1--9},
year = {2026},
publisher = {Taylor \& Francis}

}

@inproceedings{Collins2024multi,
author = {Collins, Liam and Hassani, Hamed and Soltanolkotabi, Mahdi and Mokhtari, Aryan and Shakkottai, Sanjay},
title = {Provable multi-task representation learning by two-layer ReLU neural networks},
year = {2024},
booktitle = {Proceedings of the 41st International Conference on Machine Learning},
articleno = {370},
numpages = {54}
}

\end{document}